\documentclass[]{kling/kling}

\usepackage{microtype}
\usepackage{graphicx}
\usepackage{subcaption}
\usepackage{csquotes}
\usepackage{afterpage}
\usepackage{xcolor}

\usepackage{amsmath}
\usepackage{amssymb}
\usepackage{mathtools}
\usepackage{amsthm}
\usepackage{xspace}
\usepackage{colortbl}
\usepackage{multirow}
\usepackage{enumitem}
\usepackage{wrapfig}
\usepackage{float}
\usepackage{nicefrac}
\usepackage[font=small,labelfont=bf,justification=raggedright]{caption}

\usepackage{booktabs}
\usepackage{pifont} 

\usepackage{geometry}
\usepackage{tabularx}

\usepackage{array}
\newcommand{\xmark}{\textcolor{red}{\ding{55}}}%

\newtcolorbox{observationpanel}[1][]{
  enhanced,
  breakable,
  colback=black!3,
  colframe=black!14,
  boxrule=0.5pt,
  arc=3pt,
  left=10pt,
  right=10pt,
  top=8pt,
  bottom=8pt,
  borderline west={2.2pt}{0pt}{black!55},
  fonttitle=\sffamily\bfseries,
  coltitle=black!80,
  title={Observation},
  #1
}

\newtcolorbox{insightpanel}[1][]{
  enhanced,
  breakable,
  colback=black!3,
  colframe=black!14,
  boxrule=0.5pt,
  arc=3pt,
  left=10pt,
  right=10pt,
  top=8pt,
  bottom=8pt,
  borderline west={2.2pt}{0pt}{black!75},
  fonttitle=\sffamily\bfseries,
  coltitle=black!80,
  title={Insight},
  #1
}

\usepackage{makecell}
\usepackage{titletoc}
\usepackage{fontawesome5}
\definecolor{myred}{RGB}{200,30,30}
\definecolor{myblue}{RGB}{30,150,200}
\definecolor{mygreen}{RGB}{70,170,80}
\newcommand{\name}{Beacon\xspace}
\newcommand{\graytext}[1]{\textcolor{gray}{#1}}
\newtcblisting[auto counter]{promptbox}[2][]{
  enhanced,
  breakable,
  listing only,
  colback=gray!10,
  colframe=gray!45,
  arc=3mm,
  boxrule=0.6pt,
  left=2mm,
  right=2mm,
  top=2mm,
  bottom=2mm,
  title={Prompt Box~\thetcbcounter: #2},
  coltitle=white,
  colbacktitle=black,
  fonttitle=\bfseries\footnotesize,
  boxed title style={
    enhanced,
    arc=2mm,
    boxrule=0pt,
    colframe=black,
    colback=black,
    left=1.5mm,
    right=1.5mm,
    top=0.5mm,
    bottom=0.5mm
  },
  attach boxed title to top left={
    xshift=3mm,
    yshift=-2mm
  },
  listing options={
    breakindent=0pt,
    basicstyle=\ttfamily\fontsize{8}{8}\selectfont,
    breaklines=true,
    columns=fullflexible,
    keepspaces=true,
    showstringspaces=false
  },
  #1
}

\newcommand{\MA}{\textit{Mode Adaptiveness}\xspace}
\newcommand{\TE}{\textit{Tool Effect}\xspace}
\newcommand{\MAtext}{\textit{$\text{MA}_{\text{text}}$}\xspace}
\newcommand{\MAtool}{\textit{$\text{MA}_{\text{tool}}$}\xspace}
\newcommand{\MAmean}{\textit{$\text{MA}_{\text{mean}}$}\xspace}
\newcommand{\toolgain}{\textit{$\text{Tool-Gain}$}\xspace}
\newcommand{\toolharm}{\textit{$\text{Tool-Harm}$}\xspace}

\newcommand{\textretain}{\textit{Text-Retain}\xspace}
\newcommand{\toolacc}{\textit{Tool-Available Acc.}\xspace}
\newcommand{\textacc}{\textit{Tool-Free Acc.}\xspace}

\title{Beacon: Knowing When and How to Perform Agentic Visual Reasoning}
\headertext{Beacon: Knowing When and How to Perform Agentic Visual Reasoning}

\makeatletter
\renewcommand\author[2][]{\addtolist[#1]{#2}{\authorlist}{\authorformat}{\quad}}
\newcommand{\authorbreak}[2][]{\addtolist[#1]{#2}{\authorlist}{\authorformat}{\\[0pt]}}

\renewcommand\contribution[2][]{\addtolist[#1]{#2}{\contributionlist}{\contributionformat}{\quad}}
\makeatother

\author[1,2,*]{Qixun Wang}
\author[1,2,*,\dagger]{Yang Shi}
\author[1]{Letian Cheng}
\author[1,2]{Zhuoran Zhang}
\author[1]{Yan He}
\author[3,2]{Yuqi Tang}
\author[1]{Qi Zhang}
\authorbreak[4]{Xinlei Yu}
\author[5]{Ruizhe Chen}
\author[6]{Tianrun Xu}
\author[2,\dagger]{Yuanxing Zhang}
\author[2]{Pengfei Wan}
\author[6]{Haotian Wang}
\author[1,\ddagger]{Xianghua Ying}

\affiliation[1]{Peking University}
\affiliation[2]{Kling Team}
\affiliation[3]{HKUST(GZ)}
\affiliation[4]{CUHK}
\affiliation[5]{ZJU}
\affiliation[6]{THU}
\contribution[*]{Equal Contribution}
\contribution[\dagger]{Project Lead}
\contribution[\ddagger]{Corresponding Author}

\abstract{
The fundamental goal of agentic visual reasoning is to improve the success rate of multimodal large language models (MLLMs) on complex tasks.
In this work, we rethink agentic visual reasoning through two key dimensions of tool use: \textbf{\MA} and \textbf{\TE}.
\MA characterizes whether an MLLM can recognize when tools are truly necessary and invoke them accordingly, thereby avoiding unnecessary computational overhead while improving performance on challenging problems that require tool assistance.
\TE characterizes the actual impact of tool use: tools should extend the model's capabilities on problems unsolvable through tool-free reasoning, while avoiding additional errors on problems that the model can already solve without tools.
We conduct a comprehensive analysis to quantify these two properties and empirically reveal that existing agentic visual reasoning models exhibit limited \MA, while the gains produced by tool use on hard examples are largely offset by the harm introduced on easy examples that the models can already solve.
Motivated by these observations, we propose \textbf{\name}, a novel agentic visual reasoning model trained with supervised fine-tuning (SFT) and reinforcement learning (RL). 
At the core of \name are the \textit{Necessity-Aware Adaptive Reward} and the \textit{Hint-Guided Capability Expansion} mechanism in the RL stage. \textit{Necessity-Aware Adaptive Reward} encourages tool-free solutions when they succeed while preserving full reward for successful tool use when tool-free rollouts fail. Hint-Guided Capability Expansion uses verified, answer-free expert hints to recover learning signals from all-wrong rollout groups, aiming at extending the tool-use capability on the hardest problems.
Across 13 benchmarks, Beacon achieves the highest average score among the evaluated open-source models and ranks first on 11 benchmarks. On five diagnostic benchmarks, it improves the average tool-available accuracy over its tool-free accuracy by 1.96 points and achieves the largest tool-gain minus tool-harm score (+3.14 points). These results show Beacon's advanced performance, \MA, and the net benefit of tool use.
Across 13 benchmarks, Beacon achieves the highest average score among the evaluated open-source models and ranks first on 11 benchmarks. On five diagnostic benchmarks, it improves the average tool-available accuracy over its tool-free accuracy by 1.96 points and achieves the largest tool-gain minus tool-harm score (+3.14 points). These results show Beacon's advanced performance, \MA, and the net benefit of tool use.
}

\metadata[Resources]{\faGithub\,\href{https://github.com/NOVAglow646/Beacon}{Code}\qquad
\faCube\,\href{https://huggingface.co/NOVAglow646/Beacon}{Model}}

\begin{document}
\maketitle

\begin{figure}[t]
    \centering
    \includegraphics[width=\textwidth]{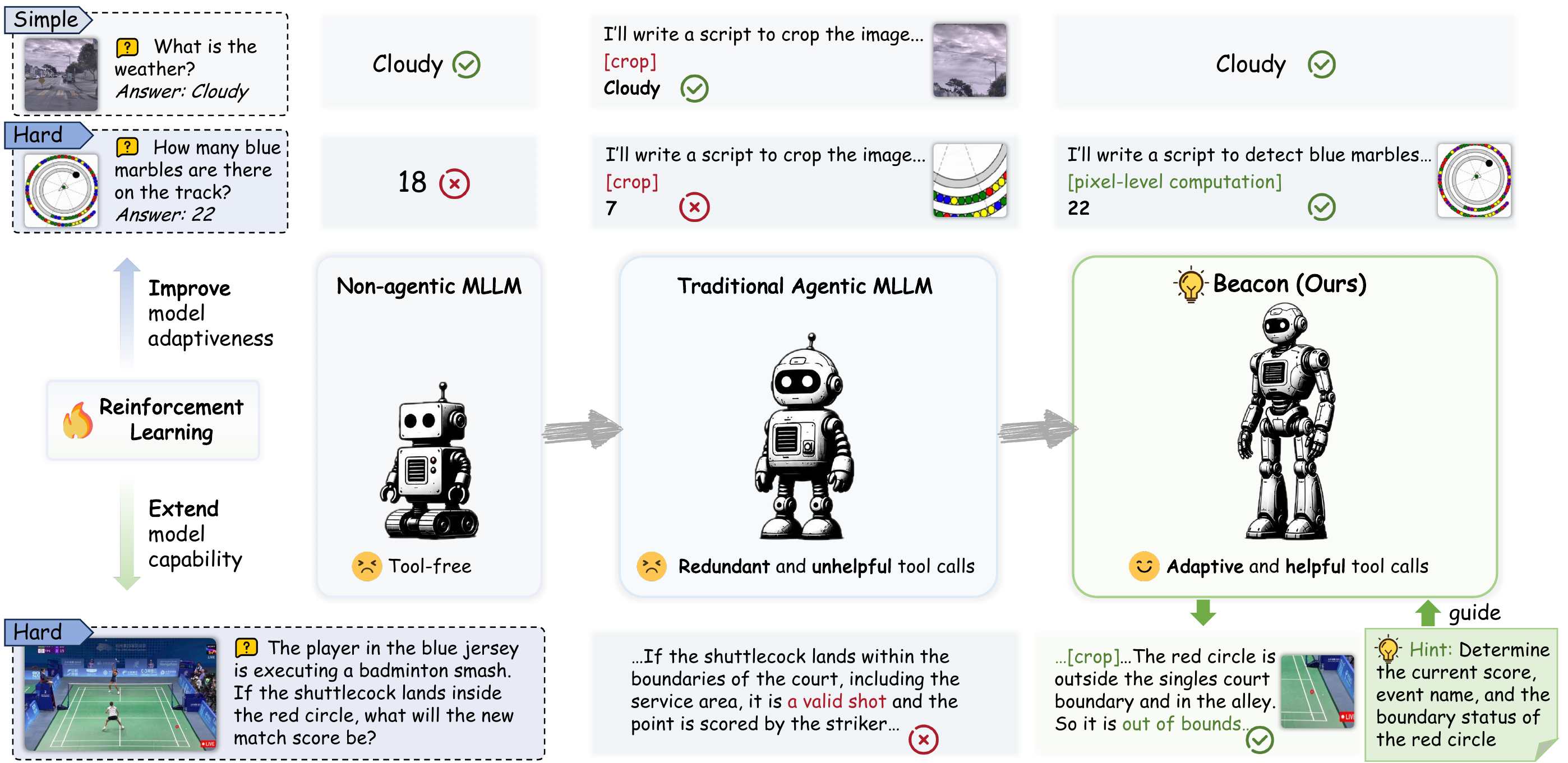}
    \caption{\textbf{Agentic visual reasoning models should use tools \textit{adaptively} and \textit{effectively}.}}
    \label{fig:overview-teaser}
\end{figure}

\begin{figure}[htbp]
    \centering

    \begin{subfigure}{0.32\linewidth}
        \centering
        \includegraphics[width=\linewidth]{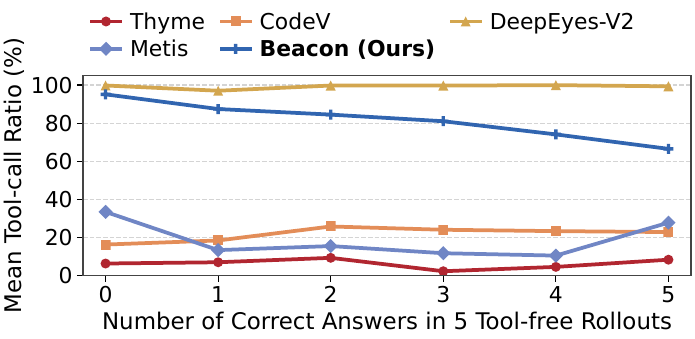}

        \caption{}
        \label{fig:teaser_a}
    \end{subfigure}
    \hspace{0.01\linewidth}
    \begin{subfigure}{0.32\linewidth}
        \centering
        \includegraphics[width=\linewidth]{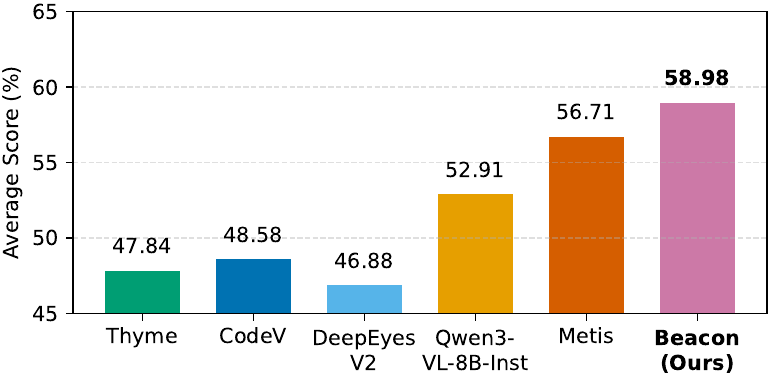}

        \caption{}
        \label{fig:teaser_b}
    \end{subfigure}
    \begin{subfigure}{0.32\linewidth}
        \centering
        \includegraphics[width=\linewidth]{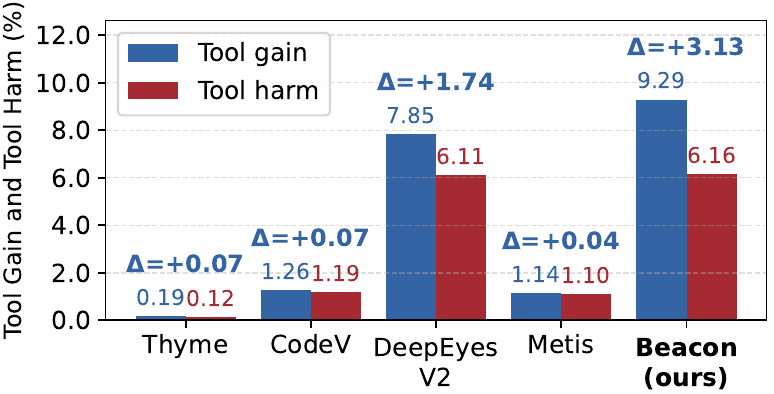}

        \caption{}
        \label{fig:teaser_c}
    \end{subfigure}

    \vspace{-0.5em}
    \caption{
    \textbf{(a)} The tool-call ratio on samples with different levels of tool-free accuracy.
    \textbf{(b)} Average performance across 13 benchmarks.
    \textbf{(c)} Tool-induced performance gains and harms (see Section~\ref{sec:metrics} for details).
    \textbf{Takeaway 1}: Beacon exhibits adaptive tool-invocation behavior that is largely absent from existing agentic visual reasoning models, invoking tools more frequently on samples that are difficult to solve through tool-free reasoning (Figure \ref{fig:teaser_a}).
    \textbf{Takeaway 2}: Beacon achieves the best average performance and the largest gap between tool gain and tool harm (Figure \ref{fig:teaser_b} and \ref{fig:teaser_c}), indicating that its tool use more effectively extends the model's capabilities beyond tool-free reasoning. In contrast, for prior models, the gains from tool use are largely offset by the errors it introduces, resulting in little improvement over tool-free reasoning.
}
    \label{fig:teaser}
\end{figure}

\section{Introduction}

Agentic visual reasoning~\citep{zhang2025thyme,yan2026metis,hong2026deepeyesv,hou2025codev}, which leverages external tools to produce intermediate multimodal results that support final-answer generation, has emerged as an important paradigm for multimodal large language models (MLLMs).
Although prior studies have demonstrated empirical gains, they often overlook two critical aspects of tool use (see Figure \ref{fig:teaser}): (1) \emph{tool-invocation adaptiveness}: whether a model can autonomously determine when tools are necessary based on factors such as tool-free solvability; and (2) \emph{tool effect}: whether tool use extends the model's capabilities on problems that cannot be solved without tools while avoiding performance degradation on problems that are solvable through tool-free reasoning.

To investigate these aspects, we conduct a comprehensive evaluation of representative agentic visual reasoning models. 
Our analysis reveals that existing models generally exhibit limited adaptiveness in tool invocation, while the gains from tool use are often offset by the errors it introduces, yielding little improvement over tool-free reasoning. 
Motivated by these findings, we propose \textbf{\name}, an agentic visual reasoning model with improved tool-invocation adaptiveness (Figure \ref{fig:teaser_a}) and stronger tool-use capability, enabling genuine gains beyond tool-free reasoning (Figure \ref{fig:teaser_c}).

During inference, \name can autonomously generate Python code to perform diverse visual and numerical operations, including image cropping, annotation, enhancement, rotation, stitching, pixel-level computation, and complex numerical calculations. To train \name, we first construct a high-quality data synthesis pipeline that equips the model with fundamental code-use capabilities. We then introduce a reinforcement learning (RL) framework consisting of a necessity-aware adaptive reward, which encourages tool use only when needed, and a hint-guided rollout strategy, which exposes the model to effective tool-use trajectories on problems that cannot be solved without tools.

Experiments on 13 benchmarks show that \name achieves the best average performance among open-source models. Further analysis demonstrates that \name exhibits stronger tool-invocation adaptiveness and the largest gap between tool-induced gain and harm, indicating that our RL framework improves both the adaptiveness and effectiveness of tool use. We compare our work with previous works in Appendix \ref{sec:related_work}.

We summarize our contributions as follows:
\begin{itemize}
    \item We analyze tool-invocation adaptiveness and the effects of tool use in representative agentic visual reasoning models. Our results reveal that existing models generally lack adaptive tool-use behavior, while the gains achieved on hard problems are largely offset by the errors introduced on problems that are already solvable through tool-free reasoning (Section~\ref{sec:metrics}).

    \item We develop a high-quality supervised fine-tuning (SFT) data synthesis pipeline, an RL data curation strategy, and a novel RL framework to train \name. These designs improve the model's tool-invocation adaptiveness and strengthen its ability to use tools to extend its capabilities beyond tool-free reasoning (Section~\ref{sec:method}).

    \item We evaluate \name on 13 benchmarks spanning diverse visual reasoning tasks. \name achieves the best average performance and exhibits stronger tool-invocation adaptiveness, together with the largest margin between tool-induced gain and harm, demonstrating that our RL framework improves both the adaptiveness and effectiveness of tool use (Section~\ref{sec:experiments}).
\end{itemize}

\section{Analysis}
In this section, we conduct comprehensive experiments to analyze two different aspects of the tool use of agentic visual models: \MA and \TE.

\subsection{The Proposed Metrics}
\label{sec:metrics}
\paragraph{\textit{Mode Adaptiveness} (\textit{MA})}
measures whether a model uses tools less when a problem can be reliably solved through tool-free reasoning, and invokes tools when tool-free reasoning is insufficient.

To assess whether a problem can be reliably solved through tool-free reasoning, we require the model to answer using Prompt~\ref{box:text_prompt} (denoted as $p_{\text{text}}$) in Appendix~\ref{app:metric_analysis_details}, which prohibits tool use. Unlike prior analyses that determine tool-free solvability from a single response~\citep{guo2026multimodal}, a procedure that can be sensitive to sampling stochasticity, we sample five tool-free responses for each problem. We classify a problem as ``text-easy'' if at least four responses are correct and as ``text-hard'' if at most one response is correct. Problems with two or three correct responses are regarded as ambiguous and excluded from the analysis. For tool-enabled reasoning, we adopt the official tool-use prompt of each model, denoted as $p_{\text{tool}}$. We minimally modify these prompts to permit, rather than mandate, tool use, thereby revealing each model's intrinsic mode-adaptation behavior.

Let $x$ denote a problem
sample and $\pi_\theta$ denote the model. Denote the ``text-easy'' and ``text-hard'' samples as $\mathcal{E}$ and $\mathcal{H}$, respectively. Let $u_{x,i}$ denote the usage of tools for the $i$-th answer of sample $x$ under the tool-enabled prompt $p_{\text{tool}}$, i.e., $u_{x,i} = 1$ if the model uses tools for the $i$-th answer of sample $x$ and $u_{x,i} = 0$ otherwise. \textit{MA} consists of the following sub-metrics.
\[
    \MAtext =
    \frac{1}{5|\mathcal{E}|}\sum_{x\in \mathcal{E}}\sum_{i=1}^5 (1-u_{x,i}),\qquad \MAtool =
    \frac{1}{5|\mathcal{H}|}\sum_{x\in \mathcal{H}}\sum_{i=1}^5 u_{x,i}.
    \]
    \MAtext measures the average ratio of tool-free responses on the ``text-easy'' samples.
    A higher value indicates that the model is better at avoiding
    unnecessary tool use.
    \MAtool measures the average ratio of tool-assisted responses on the ``text-hard'' samples.
    A higher value indicates that the model is better at invoking
    tools when tool-free reasoning fails frequently. We define $\textit{MA}_{\text{mean}}=\frac{\textit{MA}_{\text{text}}+\textit{MA}_{\text{tool}}}{2}$.

\paragraph{\textit{Tool Effect} (\textit{TE})} measures how tool use affects
the model's task-solving performance compared to tool-free reasoning, including both the gains obtained on ``text-hard''
 samples and harms
introduced on ``text-easy'' samples. Let $c_{x,i}$ denote the correctness of the $i$-th response for sample $x$, i.e., $c_{x,i} = 1$ if the model answers correctly and $c_{x,i} = 0$ otherwise. Denote the total number of test samples as $N$. \TE consists
of the following sub-metrics:
\[
\toolgain
=
\frac{1}{5N}\sum_{x\in \mathcal{H}}\sum_{i=1}^5 u_{x,i}c_{x,i},\qquad\toolharm
=
\frac{1}{5N}\sum_{x\in \mathcal{E}}\sum_{i=1}^5 u_{x,i}(1-c_{x,i}).
\]
\toolgain measures the fraction of all evaluation responses that both invoke tools and correctly answer a text-hard problem under the
tool-enabled prompt $p_{\text{tool}}$. A higher value indicates that tool use is more effective
at extending the model's problem-solving capability beyond tool-free reasoning.
\toolharm measures the fraction of all evaluation responses that invoke tools but incorrectly answer a text-easy problem under $p_{\text{tool}}$. A lower value indicates that the model is better at
avoiding performance degradation caused by ineffective
tool use. 

\paragraph{Accuracy metrics.}
We also compute several other metrics to better depict the behavior of agentic visual reasoning. To measure the overall performance of the two reasoning modes, we compute the accuracy
under the tool-enabled prompt $p_{\text{tool}}$ and the enforced-text prompt $p_{\text{text}}$ averaged over five runs, denoted as \toolacc and \textacc, respectively.

\subsection{Results and Analysis}
\label{sec:metrics_results}
\begin{figure}[H]
    \centering

    \begin{subfigure}{0.32\linewidth}
        \centering
        \includegraphics[width=\linewidth]{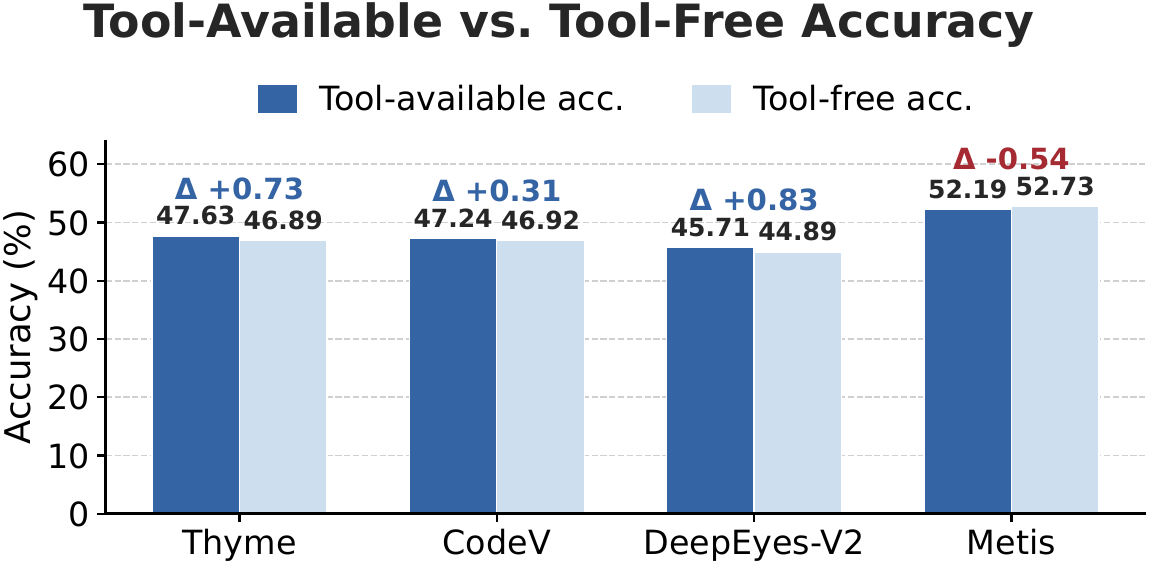}

        \caption{}
        \label{fig:motivation_a}
    \end{subfigure}
    \hspace{0.01\linewidth}
    \begin{subfigure}{0.32\linewidth}
        \centering
        \includegraphics[width=\linewidth]{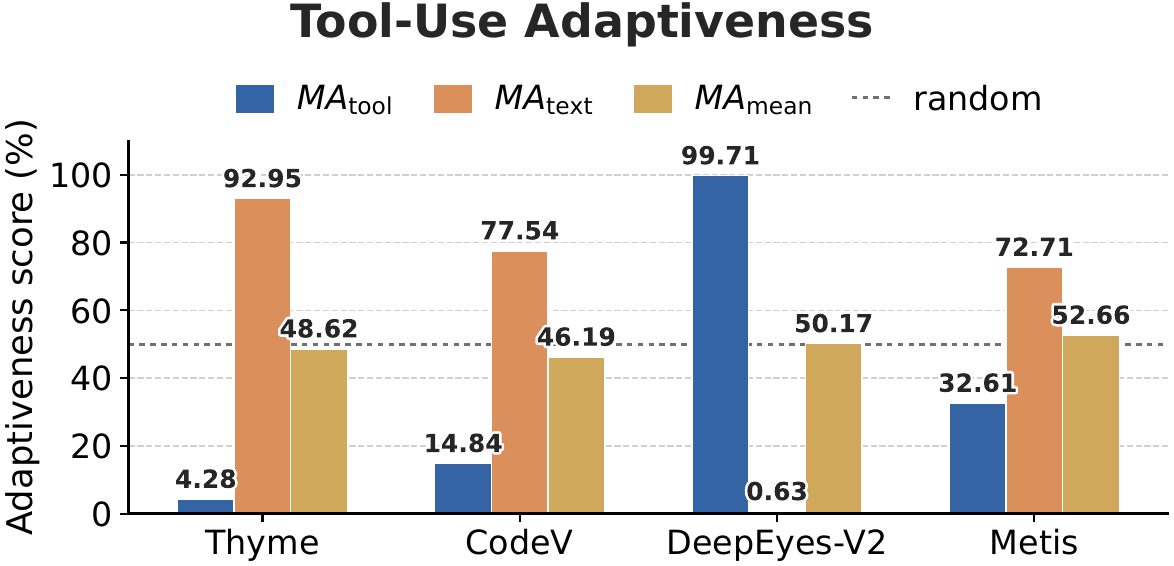}

        \caption{}
        \label{fig:motivation_b}
    \end{subfigure}
    \begin{subfigure}{0.32\linewidth}
        \centering
        \includegraphics[width=\linewidth]{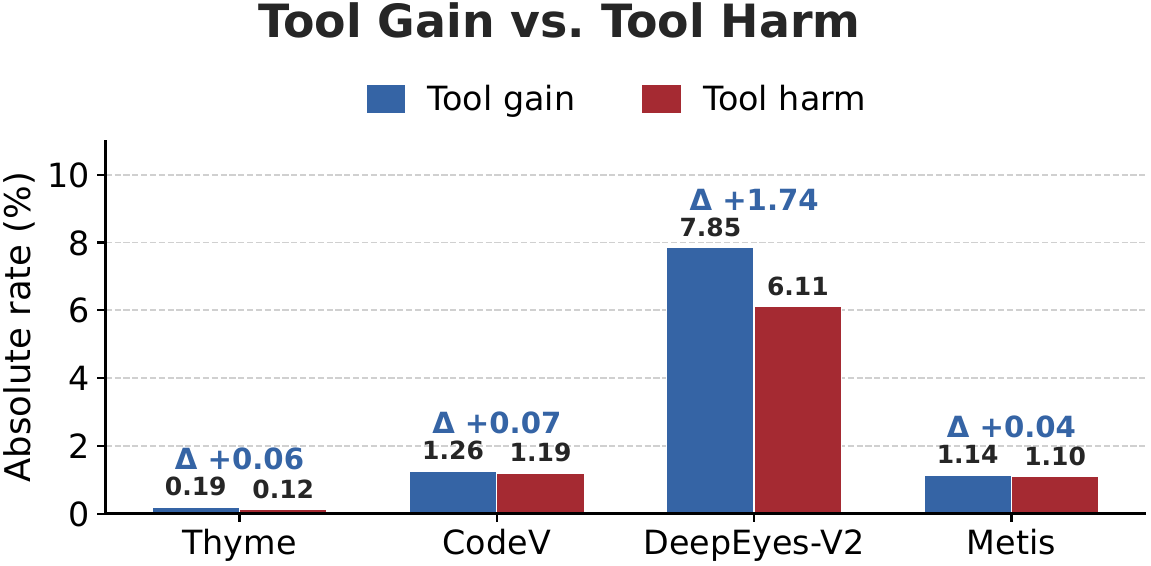}

        \caption{}
        \label{fig:motivation_c}
    \end{subfigure}

    \vspace{-0.5em}
    \caption{
    \textbf{(a)} The accuracy of tool-available and tool-free reasoning.
    \textbf{(b)} The \MAtext, \MAtool, and \MAmean of different models.
    \textbf{(c)} The \toolgain\ and \toolharm\ of different models. 
    }
    \label{fig:motivation}
\end{figure}

We evaluate several representative agentic visual reasoning models: Thyme \citep{zhang2025thyme}, DeepEyesV2 \citep{hong2026deepeyesv}, CodeV \citep{hou2025codev}, and Metis \citep{yan2026metis}, on five benchmarks: HRBench4K \citep{wang2025divide}, BLINK \citep{fu2024blink}, BabyVision \citep{chen2026babyvision}, MathVista \citep{lu2024mathvista}, and TIRBench \citep{tiri2025}. We present the overall results of the metrics averaged over these five benchmarks in Figure~\ref{fig:motivation} and summarize the key observations below. We leave the detailed results and analysis in Section \ref{sec:analysis}.

\paragraph{Observation 1 (Figure~\ref{fig:motivation_a}): Tool-enabled reasoning provides little improvement over tool-free reasoning for most existing agentic visual reasoning models.}
This finding is consistent with prior studies~\citep{guo2026multimodal,yang2026position}. By averaging results over five runs, we further consolidate this finding and show that this limited gain is robust to sampling stochasticity.
\paragraph{Observation 2 (Figure~\ref{fig:motivation_b}): Existing agentic visual reasoning models exhibit limited reasoning-mode adaptiveness.}
A model that always uses tools or never uses tools achieves an \MAmean of 50\%. Most evaluated models obtain an \MAmean close to or even below this baseline. Moreover, many consistently favor a single reasoning mode, either tool-free or tool-assisted reasoning, indicating that they fail to adapt their mode selection to the tool-free difficulty of each problem.
\paragraph{Observation 3 (Figure~\ref{fig:motivation_c}): Tool use provides limited benefit on ``text-hard'' samples for existing agentic visual reasoning models.}
Most evaluated models exhibit a low \toolgain. Although their \toolharm is also limited, \toolgain does not substantially exceed \toolharm. Consequently, tool use yields little overall performance benefit (Figure~\ref{fig:motivation_a}).

\paragraph{Hypothesis of issue attribution.}
We hypothesize that this issue arises from two primary factors: (1) the absence of an explicit objective that promotes adaptive behavior, and (2) a fundamental limitation of RL with verifiable rewards—its difficulty in expanding the model's capabilities beyond those acquired from the pretraining distribution.
\section{Method}
\label{sec:method}

\subsection{Overview}
\label{sec:overview}

Motivated by the observations in Section~\ref{sec:metrics_results}, we propose Beacon, which adopts an SFT-then-RL training paradigm. During SFT, we equip the model with fundamental code-use capabilities using synthesized trajectories spanning diverse visual reasoning tasks. During RL, we introduce adaptive reward shaping and a hint-guided rollout mechanism to further improve reasoning-mode adaptiveness and enhance the genuine performance gains brought by code use.

Given a visual query consisting of an image and a question, Beacon autonomously decides whether code execution is necessary. When code is invoked, the model generates a Python snippet enclosed in \verb|<tool_call>...</tool_call>| to manipulate the image or perform numerical computation. After execution, the resulting output is returned as an observation enclosed in \verb|<tool_response>...</tool_response>|. Then the model generates an observation about the tool results \verb|<observation>...</observation>| and continue reasoning, invoke code again, or produce the final answer.

\subsection{Training Data Construction}
\label{sec:data_construction}
\paragraph{Overview.} 
We construct a training data pool from open-source datasets. For SFT, we first evaluate the base model, Qwen3-VL-8B-Instruct~\citep{qwen3-vl}, on the source data and retain challenging examples. We then use Gemini 3.1 Pro~\citep{googledeepmind2026gemini31pro} to generate code-assisted reasoning trajectories for these examples and further refine the generated trajectories with the same model, obtaining about 15K code-assisted trajectories. For RL, we evaluate the resulting SFT model on the same data pool and select the hard examples for reinforcement learning. Statistics of the constructed training data are reported in Table~\ref{tab:dataset-statistics}, with additional details provided in Appendix~\ref{app:data_details}.

\subsection{Cold-start SFT}
We use a standard cross-entropy loss for cold-start SFT, and mask the code output (i.e., the content between \verb|<tool_response>...</tool_response>|). To avoid biasing the model towards code use and mitigate forgetting, we inject 30K correct pure-text trajectories generated by the base model for the samples with accuracy $\geq 0.6$ among five rollouts into the $\sim$15K code-assisted trajectories, obtaining $\sim$45K SFT data. See Appendix \ref{app:sft_training_details} for more details of SFT training.

\subsection{Reinforcement Learning}
\label{sec:rl}

\begin{figure}[t]
    \centering
    \includegraphics[width=0.9\linewidth]{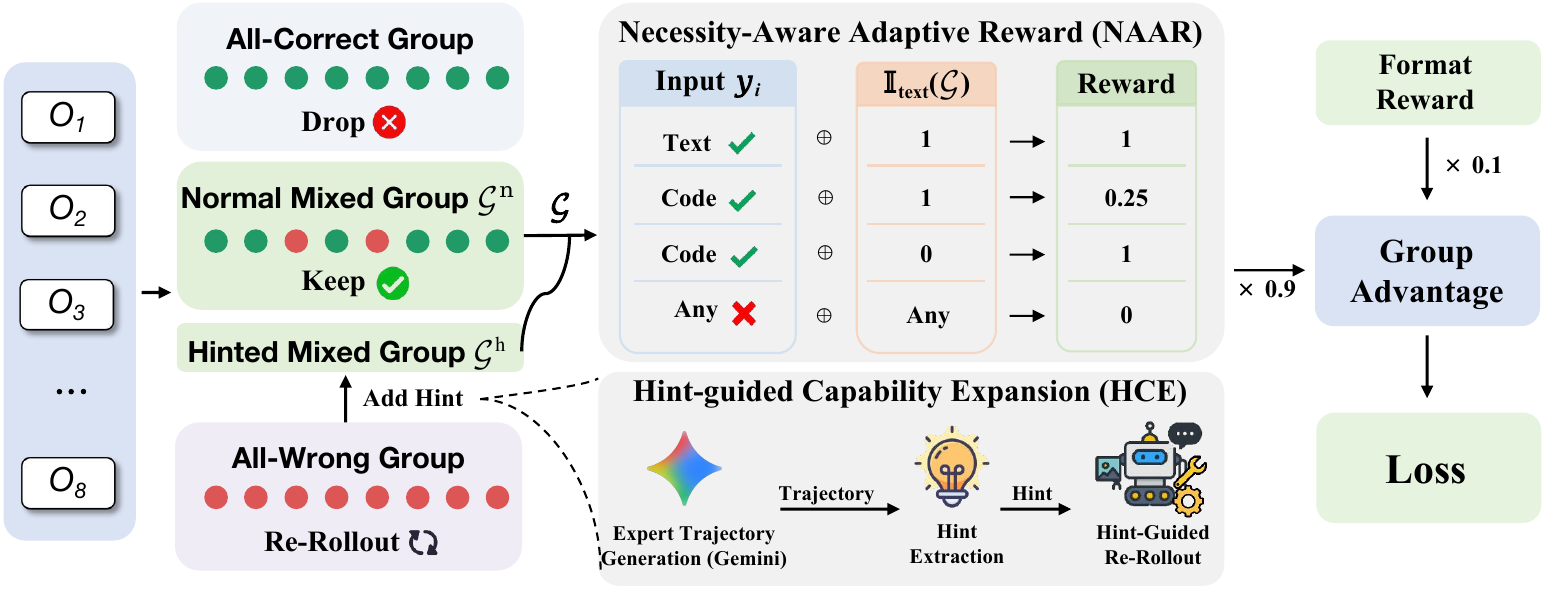}
    \caption{\textbf{Illustration of the RL process of Beacon.} The \textit{Necessity-Aware Adaptive Reward} (NAAR) and \textit{Hint-Guided Capability Expansion} (HCE) are designed to improve reasoning-mode adaptiveness and expand the tool-use capability on the most challenging problems, respectively. }
    \label{fig:RL_main}
\end{figure}

 We adopt GRPO \citep{shao2024deepseekmath} as the base RL algorithm, with our designed \emph{Necessity-Aware Adaptive Reward} and \emph{Hint-guided Capability Expansion} mechanisms to further improve the model's reasoning mode adaptiveness and the true gain of code use.

\subsubsection{Necessity-Aware Adaptive Reward}
\label{sec:adaptive_reward}
A central challenge in adaptive reasoning is to discourage unnecessary tool use without suppressing the model's ability to solve difficult problems with code. A strict binary preference for tool-free reasoning may over-penalize code-based responses even when they produce correct answers, which can introduce conflicts between the adaptive objective and the task-accuracy objective. To balance reasoning efficiency and task success, we introduce the \emph{Necessity-Aware Adaptive Reward} (NAAR), which prioritizes tool-free reasoning when it is sufficient, while still assigning partial credit to correct code-based solutions.

For each rollout group $\mathcal{G}$, let $\mathbb{I}_{\mathrm{text}}(\mathcal{G})$ indicate whether the group contains at least one correct tool-free response. The adaptive reward for a response $y_i$ is defined as
\begin{equation}
R_{\mathrm{adaptive}}(y_i;\mathcal{G}) =
\begin{cases}
1, &
\mathbb{I}_{\mathrm{text}}(\mathcal{G})=1,\;
y_i \text{ is tool-free and correct}, \\[2pt]
0.25, &
\mathbb{I}_{\mathrm{text}}(\mathcal{G})=1,\;
y_i \text{ uses code and is correct}, \\[2pt]
1, &
\mathbb{I}_{\mathrm{text}}(\mathcal{G})=0,\;
y_i \text{ uses code and is correct}, \\[2pt]
0, & \text{otherwise}.
\end{cases}
\label{eq:naar}
\end{equation}

When a problem can be solved through tool-free reasoning, NAAR assigns the highest reward to correct tool-free responses and a reduced reward to correct code-based responses, thereby encouraging the model to avoid unnecessary tool calls. Importantly, correct code-based responses are not assigned zero reward, since they still demonstrate successful reasoning and may provide useful learning signals. When no tool-free response in the group is correct, correct code-based responses receive the full reward, encouraging the model to use tools when they are necessary for solving the problem. Therefore, NAAR implements a soft preference for tool-free reasoning rather than a hard prohibition on tool use, preserving task-solving capability while improving necessity awareness.

We use the same prompt for SFT, RL rollouts, and evaluation to avoid distribution shift (see Appendix~\ref{app:sft_rl_eval_prompt} for the prompt), which instructs the model to autonomously decide whether to use code.

\subsubsection{Hint-guided Capability Expansion}
A fundamental limitation of conventional RLVR is that hard examples are often
``wasted'' when the policy fails to solve them within a limited number of
rollouts. In this case, all responses receive similarly low rewards and thus
provide little useful group-relative learning signal. Nevertheless, these hard
examples are particularly valuable for expanding the capability boundary of the
policy, which aligns with the goal of tool use. To better exploit such examples, we propose \emph{Hint-guided Capability
Expansion} (HCE), which injects expert-generated hints into the rollout process
to help the policy discover successful reasoning and tool-use trajectories on the hard examples.

Concretely, for each problem, we first sample a group
$\mathcal{G}^{\mathrm{n}}=\{y_i^{\mathrm{n}}\}_{i=1}^{N}$ from the policy using
the original prompt $x$, where $N=8$ in our experiments. If none of these
responses is correct, we regard the problem as a hard example and invoke a strong expert model
(Gemini-3.1-Pro) to construct an answer-free hint. The generated hint is then
injected into the rollout prompt, allowing the policy to explore reasoning and
tool-use behaviors that are difficult to discover from its original policy
distribution. During policy optimization, we remove the hint from the model
input while retaining the trajectories generated with the hint, thereby
transferring the hint-assisted behaviors back to the original, hint-free policy
and avoiding reliance on hints at inference time.

\paragraph{Hint generation.}
Directly prompting the expert model to produce a hint may result in unreliable
guidance for two reasons. First, the expert model may itself fail to solve the
problem correctly. Second, hint steps produced without actual execution may be impractical or unhelpful. We therefore adopt a two-stage hint-generation procedure. First, the
expert model is prompted to solve the hard example by producing a complete
code-assisted reasoning trajectory, and we retain only the correct ones. Second, the expert model extracts the
crucial reasoning steps (text or code-use steps) and the expected subgoal of each step from the verified
trajectory and converts them into an answer-free hint. Details of the
hint-generation prompts are provided in
Appendix~\ref{app:hint_generation}.

\paragraph{Hinted rollout.}
Let $h$ denote the hint generated for a hard example with original question prompt $x$.
We append the hint to the original question to construct the hinted prompt $x^{\mathrm{h}} = x \oplus h$, where $\oplus$ denotes prompt concatenation. The hint contains only intermediate
instructions, expected subgoals, and useful tool-use strategies; it does not
contain the final answer. We then sample another group of $N$ responses from
the policy model $\pi_{\theta}$ under the hinted prompt $x^{\mathrm{h}}$: $y_i^{\mathrm{h}}
\sim
\pi_{\theta}
\left(\cdot\mid x^{\mathrm{h}}\right)$ and get the hinted group $\mathcal{G}^{\mathrm{h}}
=
\left\{y_i^{\mathrm{h}}\right\}_{i=1}^{N}$.
The hinted groups $\mathcal{G}^{\mathrm{h}}$ are combined with normal groups $\mathcal{G}^{\mathrm{n}}$, i.e.,
groups sampled from the original prompt $x$ for subsequent policy optimization. For the hinted groups, when computing the policy update, we preserve the generated
trajectory $y_i^{\mathrm{h}}$ but replace $x^{\mathrm{h}}$ with the original prompt $x$ to avoid the model relying on hints at inference time.

\subsubsection{Policy Optimization with Normal and Hinted Rollouts}
Let $\mathcal{B}_{\mathrm{n}}$ and $\mathcal{B}_{\mathrm{h}}$ denote the sets of
retained responses from normal and hinted groups, respectively. For either type
of group $\mathcal{G}$, we combine the format reward $R_{\mathrm{format}}$ with the previously defined
necessity-aware adaptive reward:
\begin{equation}
R(y_i;\mathcal{G})
=
0.1R_{\mathrm{format}}(y_i)
+
0.9R_{\mathrm{adaptive}}(y_i;\mathcal{G}),
\label{eq:combined_sequence_reward}
\end{equation}
in which the format reward is defined as $R_{\mathrm{format}}(y_i)=1$ if each code block is followed by a corresponding observation block and the final answer is put in \verb|<answer>...</answer>|, and $R_{\mathrm{format}}(y_i)=0$ otherwise. 
For each group, we compute the group-relative advantage for response $y_i$ as
$A_i
=
\frac{
R(y_i;\mathcal{G})-\mu_{\mathcal{G}}
}{
\sigma_{\mathcal{G}}+1e-6
}$, 
where $\mu_{\mathcal{G}}$ and $\sigma_{\mathcal{G}}$ are the mean and standard deviation of the rewards in group $\mathcal{G}$, respectively.

Let $y_{i,t}$ be the $t$-th trainable token and let $\tau_{i,<t}$ denote the
preceding interaction history, including previous model tokens and environment
observations. We use different importance-sampling ratios for normal and hinted
groups.
For a response in a normal group, the importance-sampling ratio is $\rho_{i,t}^{\mathrm{n}}(\theta)
=
\frac{
\pi_{\theta}
\left(y_{i,t}\mid x,\tau_{i,<t}\right)
}{
\pi_{\theta_{\mathrm{old}}}
\left(y_{i,t}\mid x,\tau_{i,<t}\right)
}$,
where $\theta_{\mathrm{old}}$ denotes the model parameter before the update. 
For a hinted response, we use the following importance sampling ratio: 
$
\rho_{i,t}^{\mathrm{h}}(\theta)
=
\frac{
\pi_{\theta}
\left(y_{i,t}\mid x,\tau_{i,<t}\right)
}{
\pi_{\theta_{\mathrm{old}}}
\left(y_{i,t}\mid x^{\mathrm{h}},\tau_{i,<t}\right)
}
$.
By removing the hint from the input prompt,  we transfer the capability
obtained through hint-guided exploration into the hint-free policy. Following \cite{nath2025adaptive}, we retain the hint in the context of the old policy to mitigate off-policy issues and stabilize training.

\paragraph{Final Training Objective.}
The final actor loss follows the standard clipped GRPO objective:
\begin{equation}
\mathcal{L}_{\mathrm{GRPO}}(\theta)
=
-
\frac{1}{Z}
\sum_{b\in\{\mathrm{n},\mathrm{h}\}}
\sum_{y_i\in\mathcal{B}_b}
\sum_{t\in\mathcal{T}_i}
\min\left(
\rho_{i,t}^{b}(\theta)A_i,\,
\operatorname{clip}\left(
\rho_{i,t}^{b}(\theta),1-\epsilon,1+\epsilon
\right)A_i
\right),
\label{eq:total_hce_grpo_loss}
\end{equation}
where $\epsilon=0.2$ and
$Z=\sum_{b\in\{\mathrm{n},\mathrm{h}\}}\sum_{y_i\in\mathcal{B}_b}
|\mathcal{T}_i|$. Here, $\mathcal{T}_i$ contains only trainable tokens and
excludes tool responses.
\section{Experiments}
\label{sec:experiments}
\subsection{Experimental Setup}
\paragraph{Training details.} We take Qwen3-VL-8B-Instruct \citep{qwen3-vl} as the base model. For SFT, we train the model for 4 epochs with a peak learning rate of $1e^{-5}$. For the RL stage, we train the model for 1 epoch with a learning rate of $1e^{-6}$. See Appendix \ref{app:training_details} for more details.

\paragraph{Evaluation Benchmarks.} As \cite{yang2026position} suggested, the evaluation for agentic visual reasoning models should consider challenging benchmarks that necessitate visual tool use. Therefore, we evaluate our model on a wide range of tasks, including high-resolution visual search: V* \citep{wu2024v}, HRBench \citep{wang2025divide}, Visual Probe \citep{lai2025mini} (hard); spatial and perceptual reasoning: RealWorldQA \citep{xai2024realworldqa}, BLINK \citep{fu2024blink}, BabyVision \citep{chen2026babyvision}; quantitative and diagrammatic reasoning: ChartQAPro \citep{masry2025chartqapro}, MathVista \citep{lu2024mathvista}, MathVision \citep{wang2024measuring}, compositional and agentic visual reasoning: VisualPuzzles \citep{song2025visualpuzzles}, GameQA \citep{tong2025game}, TIRBench \citep{tiri2025}. These benchmarks go beyond simple object recognition or local perception, requiring models to perform complex visual search, spatial transformation, diagrammatic interpretation, intermediate-state imagination, etc.

\paragraph{Baselines.} We compare our model with both strong open-source and closed-source models. For open-source models, we include Qwen3-VL-8B-Instruct \citep{qwen3-vl}, PixelReasoner \citep{su2025pixel}, Thyme \citep{zhang2025thyme}, DeepEyesV2 \citep{hong2026deepeyesv}, CodeV \citep{hou2025codev}, AdaTooler-V \citep{wang2026adatoolervadaptivetooluseimages}, PyVision-RL \citep{zhao2026pyvisionrl}, and Metis \citep{yan2026metis}. For closed-source models, we include Gemini3.1-pro \citep{googledeepmind2026gemini31pro}.

\subsection{Main Results}
The evaluation results are presented in Tables~\ref{tab:perception_general} and~\ref{tab:reasoning}. Beacon-8B achieves the highest average performance among all open-source models, ranking first on 11 of the 13 benchmarks and outperforming its base model by an average of 6.07 points.

\begin{table*}[t]
    \centering
    \caption{\textbf{Comparison on high-resolution visual search, spatial, and perceptual reasoning benchmarks.} Avg. denotes the unweighted mean across the seven benchmarks and is reported only for models with complete results. \textbf{Bold} and \underline{underline} indicate the best and second-best performance among open-source models, respectively. \graytext{Gray} indicates that the result is reported in the original paper.}
    \label{tab:perception_general}
    \resizebox{\textwidth}{!}{
    \begin{tabular}{lcccccccc}
        \toprule
        \multirow{2}{*}{\textbf{Model}}
        & \multicolumn{4}{c}{\textbf{High-Resolution Visual Search}}
        & \multicolumn{3}{c}{\textbf{Spatial \& Perceptual Reasoning}}
        & \multirow{2}{*}{\textbf{Avg.}} \\
        \cmidrule(lr){2-5}
        \cmidrule(lr){6-8}
        & \textbf{V$^{*}$}
        & \textbf{HR-Bench 4K}
        & \textbf{HR-Bench 8K}
        & \textbf{VisualProbe}
        & \textbf{RealWorldQA}
        & \textbf{BLINK}
        & \textbf{BabyVision}
        & \\
        \midrule
        \rowcolor{blue!8}
        \multicolumn{9}{c}{\textit{\textbf{Closed-Source Models}}} \\
        \cmidrule(lr){1-9}
        Gemini 3.1 Pro
        & 89.01
        & 90.75
        & 86.38
        & 40.57
        & 79.22
        & 81.22
        & 48.45
        & 73.66 \\

        \midrule
        \rowcolor{teal!8}
        \multicolumn{9}{c}{\textit{\textbf{Open-Source Models}}} \\
        \cmidrule(lr){1-9}
        Qwen3-VL-8B-Instruct
        & 84.82
        & 78.13
        & 75.75
        & 37.74
        & \underline{72.29}
        & \underline{62.86}
        & 12.37
        & 60.57 \\

        Pixel-Reasoner-7B
        & 78.01
        & 71.25
        & 67.75
        & 33.96
        & 70.20
        & 55.97
        & 11.86
        & 55.57 \\

        Thyme-7B
        & 82.72
        & 77.50
        & 72.13
        & 47.17
        & 70.59
        & 54.81
        & \underline{14.43}
        & 59.91 \\

        DeepEyesV2-7B
        & 80.63
        & 75.62
        & 65.88
        & 32.08
        & 62.35
        & 54.29
        & 14.18
        & 55.00 \\

        CodeV-7B
        & 84.29
        & 75.62
        & 67.50
        & 40.57
        & 70.59
        & 58.13
        & 12.37
        & 58.44 \\

        AdaTooler-V-7B
        & \underline{\graytext{89.8}}
        & -- & -- & -- & -- & -- & -- & -- \\

        PyVision-7B
        & \graytext{88.7}
        & \graytext{78.1}
        & \graytext{74.3}
        & -- & -- & -- & -- & -- \\

        Metis-8B
        & \textbf{90.58}
        & \underline{83.50}
        & \underline{81.63}
        & \underline{48.11}
        & 71.90
        & 62.65
        & 14.18
        & \underline{64.65} \\

        \midrule
        \rowcolor{gray!10}
        Beacon-8B (Ours)
        & 89.01
        & \textbf{84.25}
        & \textbf{81.88}
        & \textbf{50.00}
        & \textbf{73.20}
        & \textbf{65.97}
        & \textbf{18.04}
        & \textbf{66.05} \\

        \rowcolor{gray!10}
        $\Delta$ over Qwen3-VL-8B-Inst.
        & \textcolor{mygreen}{+4.19}
        & \textcolor{mygreen}{+6.12}
        & \textcolor{mygreen}{+6.13}
        & \textcolor{mygreen}{+12.26}
        & \textcolor{mygreen}{+0.91}
        & \textcolor{mygreen}{+3.11}
        & \textcolor{mygreen}{+5.67}
        & \textcolor{mygreen}{+5.48} \\
        \bottomrule
    \end{tabular}
    }
\end{table*}

\begin{table*}[t]
    \centering
    \caption{\textbf{Comparison on quantitative, diagrammatic, compositional, and agentic visual reasoning benchmarks.} Averages are computed over all available reported results.}
    \label{tab:reasoning}
    \resizebox{\textwidth}{!}{
    \begin{tabular}{lccccccc}
        \toprule
        \multirow{2}{*}{\textbf{Model}}
        & \multicolumn{3}{c}{\textbf{Quantitative \& Diagrammatic Reasoning}}
        & \multicolumn{3}{c}{\textbf{Compositional \& Agentic Visual Reasoning}}
        & \multirow{2}{*}{\textbf{Avg.}} \\
        \cmidrule(lr){2-4}
        \cmidrule(lr){5-7}
        & \textbf{ChartQAPro}
        & \textbf{MathVista}
        & \textbf{MathVision}
        & \textbf{VisualPuzzles}
        & \textbf{GameQA}
        & \textbf{TIR-Bench}
        & \\
        \midrule
        \rowcolor{blue!8}
        \multicolumn{8}{c}{\textit{\textbf{Closed-Source Models}}} \\
        \cmidrule(lr){1-8}
        Gemini 3.1 Pro
        & 75.46 & 90.40 & 89.47 & 73.46 & 81.00 & 47.57 & 76.23 \\

        \midrule
        \rowcolor{teal!8}
        \multicolumn{8}{c}{\textit{\textbf{Open-Source Models}}} \\
        \cmidrule(lr){1-8}
        Qwen3-VL-8B-Instruct
        & 41.66
        & 76.40
        & \underline{53.82}
        & 36.22
        & 36.70
        & 19.01
        & 43.97 \\

        Pixel-Reasoner-7B
        & 51.03 & 70.80 & 28.09 & 33.90 & 28.10 & 16.71 & 38.11 \\

        Thyme-7B
        & 38.66
        & 68.90
        & 25.82
        & 24.14
        & 26.00
        & 19.01
        & 33.76 \\

        DeepEyesV2-7B
        & 50.92
        & 67.30
        & 27.30
        & 34.33
        & 27.50
        & 17.04
        & 37.40 \\

        CodeV-7B
        & 46.06
        & 69.80
        & 26.02
        & 34.85
        & 28.00
        & 17.70
        & 37.07 \\

        AdaTooler-V-7B
        & -- & \graytext{74.5} & -- & -- & -- & -- & --  \\

        PyVision-7B
        & -- & -- & \graytext{28.7} & -- & -- & \graytext{19.8} & -- \\

        Metis-8B
        & \underline{52.79}
        & \textbf{77.80}
        & 53.49
        & \underline{40.50}
        & \underline{38.50}
        & \underline{21.65}
        & \underline{47.46} \\

        \midrule
        \rowcolor{gray!10}
        Beacon-8B (Ours)
        & \textbf{58.48}
        & \underline{77.10}
        & \textbf{54.57}
        & \textbf{42.89}
        & \textbf{47.30}
        & \textbf{24.03}
        & \textbf{50.73} \\

        \rowcolor{gray!10}
        $\Delta$ over Qwen3-VL-8B-Inst.
        & \textcolor{mygreen}{+16.82}
        & \textcolor{mygreen}{+0.70}
        & \textcolor{mygreen}{+0.75}
        & \textcolor{mygreen}{+6.67}
        & \textcolor{mygreen}{+10.60}
        & \textcolor{mygreen}{+5.02}
        & \textcolor{mygreen}{+6.76} \\
        \bottomrule
    \end{tabular}
    }
\end{table*}

\subsection{Analysis}
\label{sec:analysis}
In this section, we present detailed results and analysis of \MA and \TE, which are evaluated on the same five benchmarks as Section \ref{sec:metrics_results}. Among samples correctly answered under $p_{\text{text}}$ and answered without tools under $p_{\text{tool}}$, we define \textretain as the fraction that remain correct under $p_{\text{tool}}$. A higher value indicates greater
robustness of tool-free reasoning across prompts. The results are shown in Table~\ref{tab:metrics}. We summarize the key observations below.

\paragraph{Observation 1: \name can further improve its already strong tool-free reasoning by using tools.} When tools are available, \name consistently outperforms tool-free reasoning across all datasets and shows the largest average performance gain $\Delta_{\mathrm{Acc}}$ (+1.96\%), while other models exhibit weaker average $\Delta_{\mathrm{Acc}}$ with tools (below +1\%). Note that the tool-free reasoning of \name still shows clear improvements over the base model (51.57\%, compared to the average accuracy of Qwen3-VL-8B-Instruct on these five datasets in Table \ref{tab:perception_general} and \ref{tab:reasoning}, which is 49.75\%), indicating that the clear performance gain of the tool use of \name does not come at the cost of weakening the tool-free reasoning capability.

\paragraph{Observation 2: \name shows the best reasoning-mode adaptiveness among evaluated models.} Although \name shows a preference for using code (the \MAtext of \name is relatively low), it exhibits the best average \MAmean, demonstrating that it can adaptively choose the appropriate reasoning mode for different problems.

\paragraph{Observation 3: \name achieves the largest gap between Tool-Gain and Tool-Harm, along with the highest \textretain.}
\name invokes tools more frequently than most baselines, achieving the highest Tool-Gain while also incurring the highest Tool-Harm. Its greater success on text-hard samples outweighs the errors in tool-assisted responses on text-easy samples, yielding the largest $\Delta_{\mathrm{TE}}$ of +3.14 percentage points. Meanwhile, its highest \textretain indicates that, when it chooses to answer without tools, it most consistently preserves correct tool-free answers across prompts.

\paragraph{Additional analysis.}
In Appendix \ref{app:fixed_text_easy_hard}, we analyze \TE on a fixed set of ``text-easy'' and ``text-hard'' samples instead of the model-dependent sets considered in Table \ref{tab:metrics}.
In Appendix \ref{app:rl_dynamics}, we also analyze the evolution of the training dynamics of the RL stage.

\begin{table}[t]
\centering
\caption{
\textbf{The metrics that reflect the \MA and \TE of agentic visual reasoning models (\%).} The definition of all metrics can be found in Section \ref{sec:metrics}. $\uparrow$ and $\downarrow$ indicate that
higher and lower values are better, respectively. 
$\Delta_{\mathrm{Acc}}=\toolacc-\textacc$ and $\Delta_{\mathrm{TE}}=\toolgain-\toolharm$. Positive and non-positive deltas are shown in \textcolor{mygreen}{green} and \textcolor{myred}{red}, respectively. Detailed results are in Appendix \ref{app:detailed_metrics}.
}
\label{tab:metrics}

\resizebox{\linewidth}{!}{
\begin{tabular}{l|lccc|ccc|cccc}
\toprule

\multirow{2}{*}{\textbf{Dataset}}
& \multirow{2}{*}{\textbf{Method}}
& \multicolumn{3}{c|}{\textbf{Accuracy}}
& \multicolumn{3}{c|}{\textbf{Mode Adaptiveness}}
& \multicolumn{4}{c}{\textbf{Tool Effects}} \\

\cmidrule(lr){3-5}
\cmidrule(lr){6-8}
\cmidrule(lr){9-12}

&
& \textbf{\toolacc $\uparrow$}
& \textbf{\textacc $\uparrow$}
& \textbf{$\Delta_{\mathrm{Acc}}$ $\uparrow$}
& \textbf{\MAtool $\uparrow$}
& \textbf{\MAtext $\uparrow$}
& \textbf{\MAmean $\uparrow$}
& \textbf{\toolgain $\uparrow$}
& \textbf{\toolharm $\downarrow$}
& \textbf{$\Delta_{\mathrm{TE}}$ $\uparrow$}
& \textbf{\textretain $\uparrow$} \\

\midrule
& Thyme
& 47.63 & 46.89 & \textcolor{mygreen}{+0.73}
& 4.28 & 92.95 & 48.62 & 0.19 & 0.12 & \textcolor{mygreen}{+0.06} & 81.43 \\
& DeepEyesV2
& 45.71 & 44.89 & \textcolor{mygreen}{+0.83}
& 99.71 & 0.63 & 50.17 & 7.85 & 6.11 & \textcolor{mygreen}{+1.74} & -- \\
& CodeV
& 47.24 & 46.92 & \textcolor{mygreen}{+0.31}
& 14.84 & 77.54 & 46.19 & 1.26 & 1.19 & \textcolor{mygreen}{+0.07} & 71.39 \\
& Metis
& 52.19 & 52.73 & \textcolor{myred}{-0.54}
& 32.61 & 72.71 & 52.66 & 1.14 & 1.10 & \textcolor{mygreen}{+0.04} & 82.98 \\
\multirow{-5}{*}{\textbf{Average}}
& \textbf{Beacon (Ours)}
& 53.53 & 51.57 & \textcolor{mygreen}{\textbf{+1.96}}
& 94.75 & 22.91 & \textbf{58.83} & 9.29 & 6.16 & \textcolor{mygreen}{\textbf{+3.13}} & \textbf{91.00} \\

\bottomrule
\end{tabular}
}
\end{table}

\subsection{Ablation Study}
We conduct ablation studies to evaluate the contribution of each component of Beacon to the overall performance, \MA, and \TE. The results are presented in Table~\ref{tab:ablation}. We observe that our full method has the best overall performance, while using the necessity-aware adaptive reward alone achieves the best \MA. The NAAR and the HCE module improve the overall \MA and \TE compared to the SFT+GRPO and SFT models, while directly conducting GRPO brings little improvement. The complete results of the ablation study are in Appendix \ref{app:ablation}.

\begin{table*}[t]
    \centering
    \caption{\textbf{Component-wise ablation study of Beacon.} ``Overall Acc.'' indicates the average accuracy across all 13 benchmarks of Table \ref{tab:perception_general} and \ref{tab:reasoning}. The other metrics are defined in Section~\ref{sec:metrics} and are averaged over the five benchmarks and five runs in Table~\ref{tab:metrics}.}
    \label{tab:ablation}
    \resizebox{0.75\textwidth}{!}{
    \begin{tabular}{@{}ccccccccc@{}}
        \toprule
        \textbf{SFT}
        & \textbf{GRPO}
        & \textbf{NAAR}
        & \textbf{HCE}
        & \textbf{Overall Acc. $\uparrow$}
        & \textbf{\toolacc $\uparrow$}
        & \textbf{\textacc $\uparrow$}
        & \textbf{\MAmean $\uparrow$}
        & \textbf{$\Delta_{\mathrm{TE}}$ $\uparrow$} \\
        \midrule
         & & & & 52.91 & -- & -- & -- & -- \\
        $\checkmark$ & & & & 56.91 & 52.02 & 51.85 & 56.20 & \textcolor{mygreen}{+1.61} \\
        $\checkmark$ & $\checkmark$ & & & 57.10 & 52.25 & 51.40 & 56.30 & \textcolor{mygreen}{+1.39} \\
        $\checkmark$ & $\checkmark$ & $\checkmark$ & & 57.75 & 53.25 & \textbf{52.24} & \textbf{59.68} & \textcolor{mygreen}{+2.54} \\
        $\checkmark$ & $\checkmark$ & & $\checkmark$ & 57.62 & 52.94 & 51.82 & 58.36 & \textcolor{mygreen}{+2.96} \\
        $\checkmark$ & $\checkmark$ & $\checkmark$ & $\checkmark$ & \textbf{58.98} & \textbf{53.53} & 51.57 & 58.83 & \textcolor{mygreen}{+3.14}\\
        \bottomrule
    \end{tabular}
    }
\end{table*}

\section{Conclusion}

In this work, we revisit agentic visual reasoning from the perspective of \MA and \TE, to characterize whether a model can select the appropriate reasoning mode for each problem and whether tool use genuinely extends its capabilities beyond tool-free reasoning. Through a systematic evaluation of representative agentic visual reasoning models, we find that existing models often exhibit a strong preference for a single reasoning mode, while the gains brought by tools are largely offset by errors introduced through ineffective tool use.
Motivated by these findings, we propose Beacon, an agentic visual reasoning model trained with a high-quality data synthesis pipeline and a tailored reinforcement learning framework. In particular, the \textit{Necessity-Aware Adaptive Reward} encourages the model to avoid unnecessary tool calls without suppressing useful tool-assisted reasoning, while Hint-guided Capability Expansion recovers informative learning signals from difficult examples that conventional RL fails to solve. Extensive experiments across 13 diverse benchmarks demonstrate that Beacon achieves strong overall performance while substantially improving reasoning-mode adaptiveness and the relative gain of tool use.


\clearpage
\bibliographystyle{kling/plainnat}
\bibliography{paper}

\clearpage

\beginappendix
\startcontents[appendix]
\printcontents[appendix]{}{1}{\section*{Appendix Contents}}
\newpage

\section{Related Work}
\label{sec:related_work}
\paragraph{Agentic Visual Reasoning.}
Recent MLLMs \citep{qwen3-vl, shi2025mavors, wang2025monet} have moved beyond static visual inputs by interleaving textual reasoning with intermediate visual interactions. Existing approaches enable models to revisit image regions, crop or edit images, generate auxiliary visual representations, or execute code for visual manipulation and numerical computation
\citep{qi2025cogcom,su2025pixel,zheng2025chainoffocus,zhu2026vtc,sarch2025grounded,su2025openthinkimglearningthinkimages,zhang2025thyme,hong2026deepeyesv,qiao2025v,song2025codedance}.

These studies have demonstrated the potential of external tools for fine-grained perception and complex multimodal reasoning.
However, most evaluations primarily report aggregate task accuracy, providing limited evidence about whether the model can appropriately choose between direct reasoning and tool-augmented reasoning, and whether tool use genuinely extends the model's capabilities beyond tool-free reasoning.

\paragraph{Adaptive and Efficient Tool Use.}
Several recent studies have recognized that indiscriminate tool invocation can introduce unnecessary computation and propose various RL strategies to encourage more adaptive tool use
\citep{song2025codedance,wang2026adatoolervadaptivetooluseimages,yan2026metis}. 

While these works provide pioneering insights, their adaptive goals may still suffer from several limitations. CodeDance \citep{song2025codedance} uses group-level accuracy to determine the optimization direction of its adaptive reward: high or low group accuracy indicates that tools should be used less or more frequently, respectively. However, this design does not condition the reward on whether a particular trajectory has actually invoked a tool. As a result, it may incorrectly penalize the tool-use behavior of a successful tool-assisted trajectory simply because it belongs to a high-accuracy group. AdaTooler-V \citep{wang2026adatoolervadaptivetooluseimages} instead labels training samples as requiring or not requiring tools according to the tool-induced performance gain of a teacher model, Qwen2.5-VL-72B \citep{Qwen2.5-VL}. Such teacher-derived supervision may suffer from distribution mismatch between the policy and the teacher, while also being constrained by the teacher's own tool-use capability. Metis \citep{yan2026metis} adopts a simpler strategy that generally encourages fewer tool calls, but this objective lacks task-dependent adaptiveness and may suppress tool use even when it is necessary for solving difficult problems. 

Our adaptive reward design (Section \ref{sec:adaptive_reward}) addresses these limitations by following two principles: (1) mode-conditioned labeling, which labels each problem as requiring or not requiring tools based on both tool-free and tool-assisted performance, and (2) online labeling, which labels each problem based on the current policy's performance rather than a fixed teacher model, mitigating off-policy issues.

\paragraph{Understanding the Actual Benefit of Visual Tools.}
A growing line of diagnostic work questions whether the gains of agentic visual reasoning genuinely arise from intermediate visual tools.
\cite{yang2026position} find that interleaved visual results contribute marginally over tool-free reasoning and that improvements mainly originate from SFT. 
\cite{ma2026what} show that tool-use RL mainly reduces errors introduced by tool use, while making limited progress in correcting problems that remain unsolved without tools.
Similarly, \cite{guo2026multimodal} find that tool-induced performance gains and harms both contribute only marginally to the overall performance.

Despite these important observations, existing analyses do not provide a unified characterization of the reasoning-mode adaptiveness and the effect of tool execution.
Moreover, they typically study tool-free solvability through a binary outcome from a single inference run, which is intrinsically unstable and could be easily perturbed by stochasticity, while obscuring how autonomous invocation behavior varies across different levels of tool-free solvability. Lastly, to the best of our knowledge, no prior work has proposed an explicit solution to improve the genuine benefit of tool use.

In contrast, our analysis framework (Section \ref{sec:metrics}) considers both adaptiveness and the effect of tool use, while providing a more robust evaluation with multiple inference runs. We also propose a novel RL framework to improve both aspects (Section \ref{sec:rl}).

\section{Experimental Details}
\subsection{Details of the Metric Analysis Experiments}
\label{app:metric_analysis_details}
\begin{promptbox}[label={box:text_prompt}]{Prompt to enforce tool-free answers}
You are a helpful assistant.

Solve the following problem step by step by reasoning directly from the provided image and question.

You must not use any external tools, code execution, Python, or search. Do not output <code> blocks or any tool-call format.

Output format must be:
<think>...</think>
<answer>...</answer>
\end{promptbox}

\subsection{Details of the Evaluation Experiments}
We build our evaluation framework on the open-source VLMEvalKit toolkit~\citep{duan2024vlmevalkit}. 
For each model, we implement the tool-use inference pipeline following its official implementation, including the prompts, tool-call format, and execution procedure. 
Each model may perform at most 20 rounds of tool calls. 
We set the decoding temperature to 0.5 for all models.

For answer evaluation, we first apply rule-based matching between the model's final prediction and the ground-truth answer. 
If the rule-based matching fails, we provide both the prediction and the ground-truth answer to Gemini 3.1 Pro~\citep{googledeepmind2026gemini31pro} or Gemini 3 Flash~\citep{google2025gemini3flash} for semantic answer judgment.

\section{Training Data Details}
\label{app:data_details}

\subsection{Overview}
In this section, we will introduce the construction pipeline of our training data.

\paragraph{1. Source data selection.}
We select source datasets according to three principles: \textbf{1) Diversity:} the data should span a broad range of tasks and domains to promote generalization; \textbf{2) Quality:} the ground-truth annotations should be reliable and minimally noisy; and \textbf{3) Difficulty:} the tasks should be sufficiently challenging to benefit from tool use and expand the model's capabilities. Following these principles, we collect data from 12 datasets, covering real-world perception, chart understanding, OCR, STEM, spatial reasoning, and real-world agentic reasoning. Most of these datasets contain human-annotated labels. To prevent evaluation contamination, we remove all samples that overlap with the test sets used in our experiments.

\paragraph{2. SFT data synthesis.}
We construct the SFT data in three stages. (1) We sample five responses from the base model, Qwen3-VL-8B-Instruct, and retain examples that are answered correctly at most twice. (2) For these hard examples, we use Gemini 3.1 Pro~\citep{googledeepmind2026gemini31pro} to generate code-assisted reasoning trajectories containing Python code and the corresponding execution outputs, retaining only trajectories that yield correct answers. (3) Because the generated trajectories may still contain low-quality reasoning or tool-use steps, we introduce an additional refinement procedure using Gemini 3.1 Pro.

\paragraph{3. RL data construction.}
We evaluate the SFT model on the same data pool and retain the examples that are answered correctly no more than three times out of five attempts. These hard examples constitute the RL training set, on which the model is further optimized using reward signals that account for both answer correctness and the effectiveness of tool use.

\begin{table}[htbp]
\centering
\caption{\textbf{Training data statistics.} We report both the total number of raw samples and the number of samples retained after data filtering. The final retained SFT data contains 30K additional trajectories of problems that are correctly answered by the base model Qwen3-VL-8B-Instruct.}
\label{tab:dataset-statistics}
\resizebox{0.5\linewidth}{!}{
\begin{tabular}{lcccc}
\toprule
\textbf{Data Type} & \textbf{Original} & \textbf{Filtered} & \textbf{Final retained} \\
\midrule
SFT Data & 212,353 & 15,705 & 45,705 \\
RL Data & 45,886 & 15,709 & 15,709 \\
\bottomrule
\end{tabular}
}
\end{table}

Table~\ref{tab:dataset-statistics} summarizes the statistics of the training data.

\subsection{Data Sources}
\label{app:data_sources}
\name leverages a diverse collection of publicly available multimodal data sources spanning mathematical reasoning, chart understanding, embodied interaction, spatial perception, game reasoning, and general visual reasoning.

Specifically, the training data are collected from Geometry3K~\citep{lu2021inter}, OlympiadBench~\citep{he2024olympiadbench}, AgentVista~\citep{su2026agentvista}, MuirBench~\citep{wang2025muirbench}, HRScene~\citep{zhang2025hrscene}, CV-Bench~\citep{tong2024cambrian}, MMMU~\citep{yue2024mmmu}, and Vero~\citep{sarch2026vero}. 
These data sources cover a broad range of task formats and visual reasoning requirements, enabling \name to expose models to heterogeneous scenarios with different levels of tool dependency.

Mathematical and geometry-oriented sources, including Geometry3K~\citep{lu2021inter} and OlympiadBench~\citep{he2024olympiadbench}, provide training samples that require visual mathematical reasoning, geometric diagram understanding, symbolic manipulation, and multi-step scientific problem solving.
HRScene~\citep{zhang2025hrscene} and CV-Bench~\citep{tong2024cambrian} enrich the training corpus with high-resolution image understanding, region-sensitive perception, object counting, spatial relation reasoning, and 2D/3D visual understanding.
AgentVista~\citep{su2026agentvista} provides tool-oriented and agentic visual reasoning data, where models are encouraged to perform long-horizon reasoning with intermediate visual operations.
Finally, MMMU~\citep{yue2024mmmu}, MuirBench~\citep{wang2025muirbench}, and Vero~\citep{sarch2026vero} introduce broader multimodal reasoning scenarios across academic, multi-image, and general visual reasoning tasks.

\begin{figure}[t]
    \centering
    \includegraphics[width=0.6\linewidth]{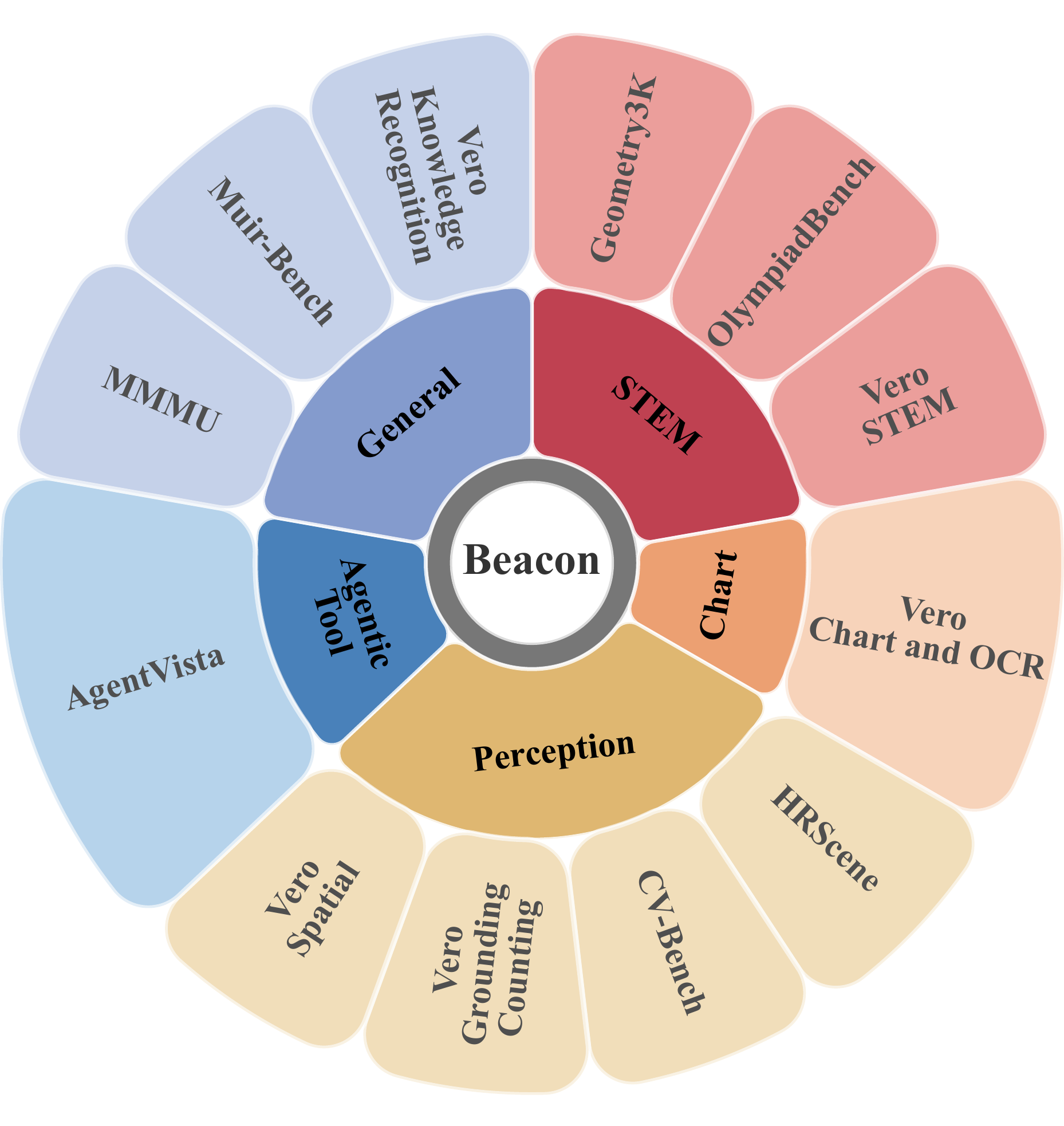}
    \vspace{-4mm}
    \caption{
    \textbf{Training data sources of \name.}}
    \label{fig:data_source}
\end{figure}

\begin{figure}[t]
    \centering
    \includegraphics[width=\linewidth]{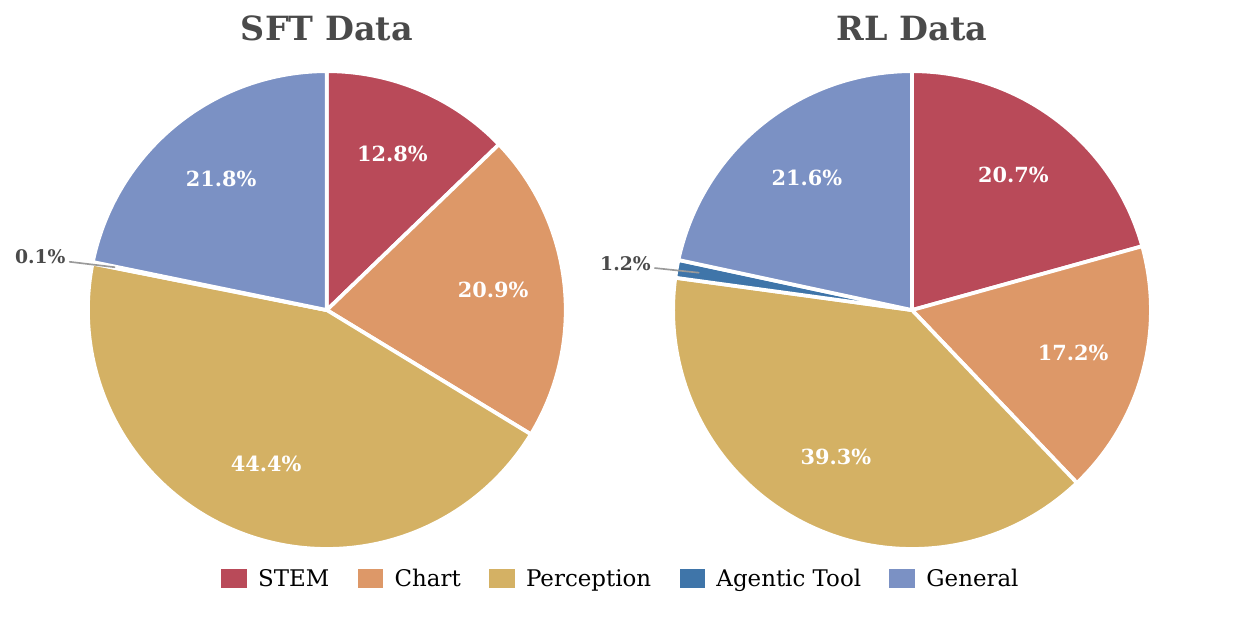}
    \vspace{-8mm}
    \caption{\textbf{Task category distribution of the SFT and RL data.}}
    \label{fig:data_sft_rl_category_distribution}
\end{figure}

\begin{figure}[t]
    \centering
    \includegraphics[width=\linewidth]{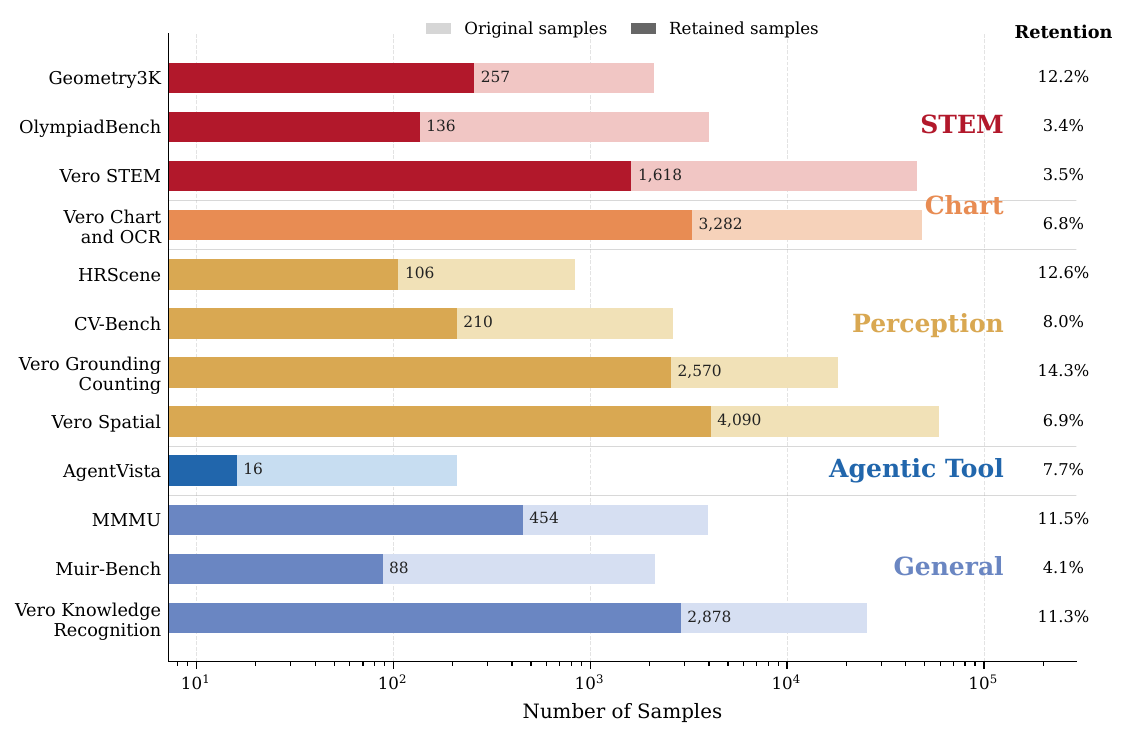}
    \vspace{-8mm}
    \caption{\textbf{Detailed statistics of the SFT data.} The numbers in the bars indicate the number of samples retained after rejection sampling, which removes easy samples, and refinement, which removes code trajectories where code does not actually help to derive the answer. “Retention” denotes the percentage of retained samples relative to the original samples.}
    \label{fig:data_details_SFT}
\end{figure}

\begin{figure}[t]
    \centering
    \includegraphics[width=\linewidth]{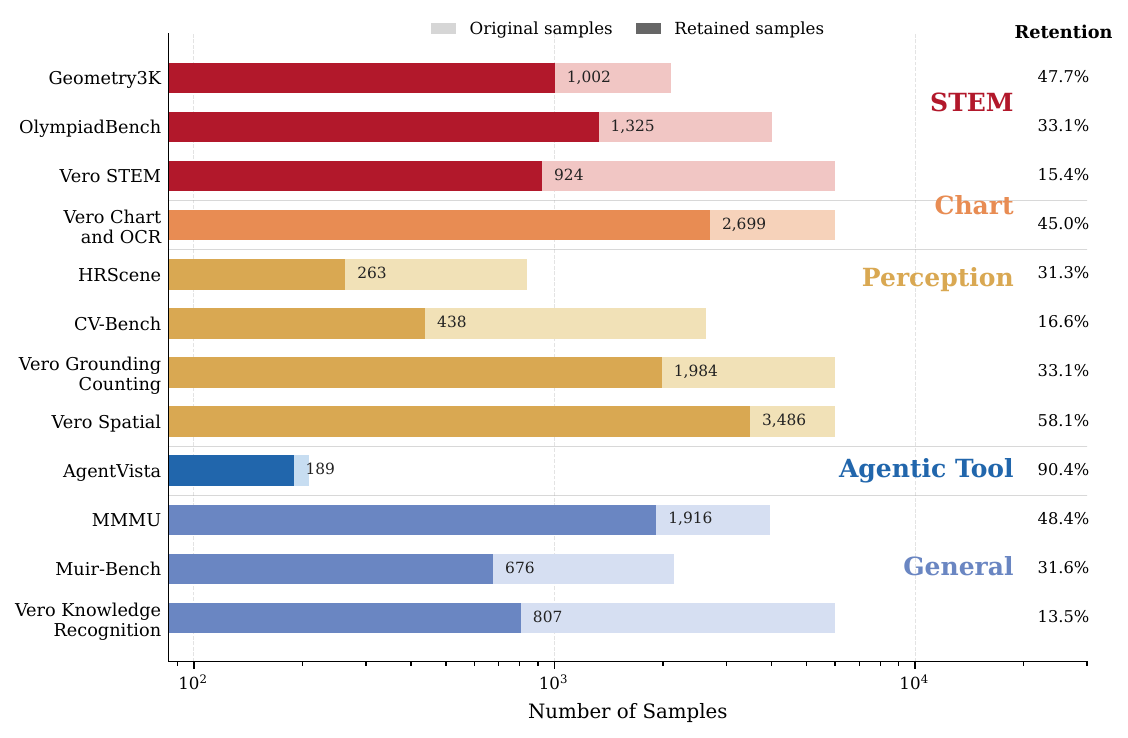}
    \vspace{-8mm}
    \caption{\textbf{Detailed statistics of the RL data.} The numbers in the bars indicate the number of samples retained after rejection sampling, which removes easy samples. “Retention” denotes the percentage of retained samples relative to the original samples.}
    \label{fig:data_details_RL}
\end{figure}

As shown in Figure~\ref{fig:data_source}, the resulting corpus spans a broad spectrum of task types, reasoning complexity, and tool requirements. 
While some samples can be solved through direct multimodal reasoning, others inherently benefit from external tools for operations such as visual manipulation, geometric analysis, or structured information extraction. 
This diversity exposes models to both tool-necessary and tool-optional scenarios, providing a foundation for learning not only how to use tools effectively, but also when tool invocation is genuinely beneficial. 
We present the task distribution of the training data in Figure \ref{fig:data_sft_rl_category_distribution}. 
Figures \ref{fig:data_details_SFT} and \ref{fig:data_details_RL} further show the number of samples from each dataset in the SFT and RL data before and after filtering, respectively.

\subsection{SFT Data Synthesis}
\label{app:sft_data_synthesis}

The synthesis of the SFT data consists of two steps: 1) generating code-based reasoning trajectories for the source problems with Gemini 3.1 Pro~\citep{googledeepmind2026gemini31pro} and retaining only those whose final answers match the ground-truth labels; and 2) refining the retained trajectories using the same model.

\paragraph{1. Prompt for SFT code-reasoning trajectory synthesis.}
The prompt is provided in Box~\ref{box:sft_data_synthesis_prompt}. Its key design choices are summarized below:

\begin{itemize}

\item \textbf{Representative code actions.}
We provide five representative code actions, including \texttt{crop}, \texttt{draw\_line}, \texttt{draw\_box}, \texttt{numeric\_calculation}, and \texttt{rotation}, covering common visual reasoning operations.
For each action, we specify its purpose, recommended library, required imports, input convention, and a concrete example.
These examples serve as guidance rather than constraints; in practice, Gemini 3.1 Pro~\citep{googledeepmind2026gemini31pro} generates a much wider variety of code patterns beyond the provided examples.

\item \textbf{Mandatory code use.}
We explicitly instruct Gemini 3.1 Pro~\citep{googledeepmind2026gemini31pro} to invoke code for obtaining visual evidence or other useful intermediate information.
Since the model may still directly answer some questions without using code, we discard trajectories that do not contain valid tool execution.

\item \textbf{Structured trajectory format.}
We require the model to generate trajectories in a structured format, using \verb|<tool_call>|...\verb|</tool_call>| for code execution, \verb|<tool_response>|...\verb|</tool_response>| for execution results, \verb|<observation>|...\verb|</observation>| for interpreting the returned evidence, and \verb|<answer>|...\verb|</answer>| for the final prediction.
This structured format improves interpretability, enables reliable automatic parsing, and encourages explicit reasoning over tool outputs before producing the final answer.

\item \textbf{Sandbox protocols.}
We define several execution protocols to reduce ambiguity and simplify parsing.
Question images and code-generated images are assigned unique identifiers (e.g., \texttt{q0}, \texttt{q1}, \texttt{g0}, and \texttt{g1}), which can be accessed through \verb|get_image(ref)|.
The final line within each \verb|<tool_call>|...\verb|</tool_call>| block must be the executable expression, returning either a PIL image object or a numeric value.
In addition, the model is instructed to obtain image dimensions via \texttt{img.size}, ensuring that coordinate-based operations are grounded in the actual image resolution.

\end{itemize}

\begin{promptbox}[label={box:sft_data_synthesis_prompt}]{Prompt for SFT Data Synthesis}
You are an image reasoning assistant. You may write Python code to obtain visual evidence or any helpful information before answering. This includes but is not limited to zooming in, drawing auxiliary lines, rotating, adjusting contrast, computing statistics, etc. Put the executable Python code inside <tool_call></tool_call> tags. When you give the final answer, you must wrap it inside <answer></answer> tags.

# Code Actions

Here are representative code patterns for acting on images and numbers. They are recommended starting points, not an exhaustive list. You may use other valid Python operations, library calls, or parameter choices as long as they fit the runtime constraints.

## crop
Purpose: Crop a rectangular region from an image using Pillow.

Recommended library: Pillow

Required imports:
from PIL import Image

Input convention:
- Read the source image from get_image("q0") / get_image("g0") or from images[index].
- Use a 4-tuple (left, top, right, bottom) as the crop box.
- The last line should be the cropped PIL image object.
Example:
<tool_call>
img = get_image("q0")
cropped = img.crop((120, 180, 420, 620))
cropped
</tool_call>

## draw_line
Purpose: Draw one or more auxiliary line segments on an image using Pillow.

Recommended library: Pillow

Required imports:
from PIL import ImageDraw

Input convention:
- Read the source image from get_image("q0") / get_image("g0") or from images[index].
- Always call copy() before drawing.
- Use ImageDraw.Draw(annotated).line((x1, y1, x2, y2), fill="red", width=4).
- The last line should be the annotated PIL image object.
Example:
<tool_call>
from PIL import ImageDraw
annotated = get_image("q0").copy()
draw = ImageDraw.Draw(annotated)
draw.line((118, 214, 612, 248), fill="red", width=4)
draw.line((132, 486, 628, 520), fill="red", width=4)
annotated
</tool_call>

## draw_box
Purpose: Draw one or more bounding boxes on an image using Pillow.

Recommended library: Pillow

Required imports:
from PIL import ImageDraw

Input convention:
- Read the source image from get_image("q0") / get_image("g0") or from images[index].
- Always call copy() before drawing.
- Use ImageDraw.Draw(annotated).rectangle((x1, y1, x2, y2), outline="red", width=3).
- The last line should be the annotated PIL image object.
Example:
<tool_call>
from PIL import ImageDraw
annotated = get_image("q0").copy()
draw = ImageDraw.Draw(annotated)
for box in [(42, 38, 128, 124), (150, 38, 236, 124)]:
    draw.rectangle(box, outline="red", width=3)
annotated
</tool_call>

## numeric_calculation
Purpose: Evaluate a mathematical expression using Python and math, instead of mental arithmetic.

Recommended library: math

Required imports:
import math

Input convention:
- Write the calculation directly in Python, such as result = 5 / 8 * 100 or result = math.sqrt(2) * 10.
- Import math only when you actually need math functions or constants.
- The last line should be the numeric result itself.
Example:
<tool_call>
import math
result = math.sqrt(2) * 10
result
</tool_call>

## rotation
Purpose: Rotate an image using Pillow to correct orientation or inspect a different view.

Recommended library: Pillow

Required imports:
from PIL import Image

Input convention:
- Read the source image from get_image("q0") / get_image("g0") or from images[index].
- Use img.rotate(angle, expand=True).
- Positive angles rotate counterclockwise.
- The last line should be the rotated PIL image object.
Example:
<tool_call>
img = get_image("q0")
rotated = img.rotate(90, expand=True)
rotated
</tool_call>

Additional reminders:

1. Even if you can already guess or know the answer from the image, you should still use sandboxed Python snippets to mark, isolate, or verify the relevant visual evidence before answering.
2. Use get_image(ref) to access runtime images. Original question images use stable ids like q0, q1, and images returned by earlier code calls use stable ids like g0, g1.
3. If you need to draw on an image, call copy() first so you do not modify the source image directly. For example: annotated = get_image("q0").copy().
4. If the current call should return an image result, the last line inside <tool_call> should be the PIL image object itself, such as cropped, rotated, or annotated. If the current call should return a numeric result, the last line should be the numeric object itself, such as result.
5. Do not do arithmetic mentally. Use Python code. You may import math when needed.
6. <tool_response>...</tool_response> is inserted by the system after the code finishes running. It is not something you should generate yourself. Never output <tool_response> tags on your own.
7. After each code execution returns a result, you must immediately output an <observation>...</observation> tag that explicitly analyzes the returned result before deciding whether to call code again.
8. The order must be strictly <tool_call>...</tool_call> -> <tool_response>...</tool_response> -> <observation>...</observation>. After that, you may continue with additional <tool_call> -> <tool_response> -> <observation> cycles.
9. Your final answer must be grounded in the tool results and observations you obtained. If a tool result is insufficient, unclear, or unhelpful, explicitly say so in <observation> and call code again to get better evidence instead of relying on your own unaided visual guess.

Full example:
Question: Is the man phone both blue and closed?
Assistant:
<tool_call>
img = get_image("q0")
img.size
</tool_call>
[The system then inserts a tool result wrapped in <tool_response>...</tool_response>]
<observation>I first checked the image size so that later coordinates are grounded in the actual canvas. I still need localized visual evidence for the phone itself.</observation>
<tool_call>
from PIL import ImageDraw
annotated = get_image("q0").copy()
draw = ImageDraw.Draw(annotated)
draw.rectangle((600, 290, 660, 380), outline="red", width=3)
annotated
</tool_call>
[The system then inserts a tool result wrapped in <tool_response>...</tool_response>]
<observation>The box covers the phone in the man's hand. The phone is visibly blue, and because he is actively holding it to his ear in use, it is not closed.</observation>
<answer>No</answer>
\end{promptbox}

\paragraph{2. Prompt for refining synthesized code-reasoning trajectories.}
Although Gemini 3.1 Pro~\citep{googledeepmind2026gemini31pro} can synthesize high-quality code-reasoning trajectories using the prompt in Box~\ref{box:sft_data_synthesis_prompt}, the generated trajectories still exhibit several common issues:

\begin{itemize}

\item \xmark~\textbf{Redundant or repetitive tool calls.}
The model may repeatedly invoke similar code operations or generate unnecessary intermediate results, resulting in unnecessarily long and inefficient trajectories.

\item \xmark~\textbf{Insufficient reliance on tool outputs.}
Although the model executes code to obtain visual evidence, the returned results are sometimes only weakly utilized. Consequently, the final answer is still primarily based on the model's direct visual reasoning rather than evidence obtained through tool execution.

\item \xmark~\textbf{Uninformative reasoning.}
Some trajectories contain repetitive or low-quality reasoning loops, where the model repeatedly produces similar thoughts or observations without making meaningful progress toward the final answer.

\item \xmark~\textbf{Missing observations of tool responses.}
The model may omit explicit observations after tool execution, making it unclear how the returned evidence is interpreted or how it supports the final prediction.

\end{itemize}

To address these issues, we further refine the synthesized trajectories with Gemini 3.1 Pro~\citep{googledeepmind2026gemini31pro} using the prompt in Box~\ref{box:sft_data_refine_prompt}. 
The refinement aims to remove trajectories whose final answers bypass tool outputs, eliminate redundant or uninformative code calls and reasoning, and enforce a cleaner trajectory structure. 
Manual inspection shows that the refined trajectories are substantially higher in quality, featuring more concise code execution, more informative reasoning, and stricter adherence to the prescribed format. 
These observations highlight the importance of trajectory refinement for high-quality automatic data synthesis.

\begin{promptbox}[label={box:sft_data_refine_prompt}]{Prompt for SFT Data Refinement}
You are refining a visual reasoning trajectory with tool calls.
Do not call tools again. Do not invent tool outputs. Only refine the trajectory text while keeping the final answer unchanged.
You will see the original question first, then the trajectory messages after the question.
The trajectory uses the following conventions:
1. Tool returns are wrapped in <tool_response>...</tool_response> and images are inserted in place.
2. <tool_call>...</tool_call> is executable code.
3. <observation>...</observation> is the observation after a tool returns.
4. <answer>...</answer> is the final answer.
5. Text thoughts may be interleaved anywhere in the trajectory.
Your task:
1. Check whether the trajectory actually relied on useful tool information to answer the question. If the tools did not provide helpful information and the answer was effectively derived from tool-free reasoning, especially in the last step, discard the trajectory.
2. Check whether the same tool was called repeatedly with unhelpful retries. Remove the unhelpful repeated tool-call steps and keep only the step that produced the key useful information.
3. For each tool call, if the tool result is followed by no meaningful <observation>, add one.
4. Remove clearly meaningless tool calls or meaningless code, such as trivial assignments, like "<tool_call>result = 1result</tool_call>".
5. Remove meaningless thought fragments.
6. Note that `img.size` is necessary for the model to know the image size, so keep the `img.size` tool call that corresponds to it even if it looks trivial.
7. Check whether each <observation> block precisely describes the preceding tool call result. If not, try to refine the <observation> to make it more accurate and specific to the actual tool response.
Output requirements:
1. Output JSON only.
2. If no refinement is needed, output {"decision":"keep"}.
3. If the trajectory should be discarded, output {"decision":"drop","reason":"..."}.
4. If refinement is needed, output JSON in this schema:
{
  "decision": "refine",
  "messages": [
    {"kind": "assistant", "text": "... assistant message text ..."},
    {"kind": "original_message", "source_message_id": "M3"}
  ]
}
5. In "messages", do not repeat the question or system prompt. Only output the messages after the original question.
6. Use {"kind":"original_message","source_message_id":"Mx"} only for original non-assistant messages that should be kept exactly, especially tool-response user messages.
7. For assistant messages, put the full assistant text in "text". If a tool-response step is kept, keep its <tool_response> in the corresponding original_message instead of rewriting it.
8. Keep the final answer unchanged.
9. If a <tool_call> step is removed, also remove the corresponding <tool_response> and <observation>.
\end{promptbox}

\section{Training Details}
\label{app:training_details}

\subsection{Details of Cold-start SFT}
\label{app:sft_training_details} 
For SFT, we train Qwen3-VL-8B-Instruct for 4 epochs using Megatron~\citep{megatron-lm} on the ms-swift framework~\citep{zhao2024swiftascalablelightweightinfrastructure}. 
We optimize the model with AdamW \citep{loshchilov2017decoupled} using a peak learning rate of $1 \times 10^{-5}$, a weight decay of 0.1, and a cosine learning-rate schedule with a 5\% warm-up ratio. 
Gradients are accumulated over 128 micro-batches before each parameter update.

\subsection{Details of RL}
We use VeRL~\citep{sheng2024hybridflow} as the RL training framework. 
RL training is conducted for 1 epoch on 64 NVIDIA H200 GPUs. 
For each prompt, the policy generates 8 rollouts, with a batch size of 128 prompt groups and a PPO mini-batch size of 128. 
We use an actor learning rate of \(1\times10^{-6}\) and the vanilla token-mean PPO objective with symmetric clipping bounds of 0.2, while disabling both KL and entropy regularization. 
The maximum sequence length is 30{,}720 tokens, including up to 20{,}480 prompt tokens and 10{,}240 response tokens, with at most 12 tool calls per rollout. 
For all-wrong prompt groups, structured hints are generated using up to three expert attempts and 15 expert tool calls with 32 parallel workers. If the expert fails to generate a correct solution, we simply discard this sample.
Training is performed in \texttt{bfloat16} precision with gradient checkpointing and a frozen vision encoder.

\subsection{System Prompts for SFT, RL, and Evaluation}
\label{app:sft_rl_eval_prompt}
To avoid distribution mismatch, we use the same system prompt for SFT, RL training, and evaluation. 
The prompt is shown in Box~\ref{box:sft_rl_prompt}. 
It is identical to the system prompt used for SFT data synthesis (Box~\ref{box:sft_data_synthesis_prompt}), except that the mandatory code-use instruction is replaced with an optional one (\verb|"You should decide whether to use code..."|).
\begin{promptbox}[label={box:sft_rl_prompt}]{The system prompt for SFT, RL, and evaluation}
...the same as the system prompt used for SFT data synthesis...

Additional reminders:

1. You should decide whether to use code to solve the problem. Only write codes when necessary. If you can directly solve the problem, don't use code.
2. ...

...the same as the system prompt used for SFT data synthesis...

\end{promptbox}

\subsection{Details for Hint Generation}
\label{app:hint_generation}

The system prompt for hint generation is shown in Box~\ref{box:hint_generation}. 
The generated hint consists of a sequence of reasoning steps, each containing an instruction and an expected subgoal. 
The instruction specifies the reasoning, observation, or tool action to perform, while the subgoal describes the intermediate evidence or outcome to be obtained without revealing the final answer. 
In our current implementation, we retain only the \verb|"hint"| field as the training hint. 
The remaining fields, \verb|"tool_steps"| and \verb|"subgoals"|, are preserved for analysis and may serve as fine-grained supervision or reward signals in future work.

After obtaining the hinted group, we drop the original group with all-wrong responses. 

\begin{promptbox}[label={box:hint_generation}]{The system prompt for hint generation}
  
You will convert a correct multimodal reasoning trajectory into a hint for a weaker policy model.

The hint must NOT include the final answer, the reference answer, or an explicit restatement of either answer. It should describe the useful method, evidence to seek, and tool-use plan only.

First identify which steps in the Gemini trajectory are truly necessary for reaching the answer, then condense only those key steps into a short plan. Do NOT copy the full trajectory, every tool call, or incidental observations. Keep only the reasoning, visual observation, or tool steps that are most useful for solving the task.

The hint field itself must be a numbered list. Each item must use exactly this pattern: "1. [Instruction] xxx; [Expected Subgoal] xxx;". The Instruction should state the reasoning, observation, or tool action to perform. The Expected Subgoal should state what intermediate evidence or result that step should obtain, without revealing the final answer.

Prefer actionable tool steps over vague advice. Describe code actions at a high level rather than copying large code blocks from the trajectory. The tool field should be a concise semantic operation label derived from the Python code content, not merely the word python. Use exact canonical labels from the code-action prompt when applicable: crop, draw_line, draw_box, numeric_calculation, rotation.

For each tool step, subgoal_result must summarize the actual evidence or result obtained after the code ran. Do not write imperative sentences such as 'crop the image' or 'check the labels' in subgoal_result; write factual statements such as 'the crop shows ...' or 'the calculation returns ...'. If a tool result contains exact numeric values, coordinates, counts, OCR strings, table cells, or other precise evidence, include them explicitly in subgoal_result. Make subgoal_result as informative as needed to show what was learned and why it matters for that step's subgoal, while keeping it answer-free.

Extraction mode:
Extract the critical tool-use sequence and, for each step, state what information that tool step was meant to obtain. Omit tool calls or observations that did not help. If the trajectory did not use tools, extract the key reasoning idea instead.

Subgoal instruction:
Return subgoals as exactly the ordered list of subgoal_result strings from tool_steps. This keeps the training-facing subgoals grounded in actual tool results rather than intended actions.

Return exactly one JSON object with this schema:
{
  "hint": "1. [Instruction] key reasoning, observation, or tool step; [Expected Subgoal] answer-free intermediate evidence or result;\n2. [Instruction] next key step; [Expected Subgoal] next answer-free intermediate evidence or result;",
  "tool_steps": [
    {
      "step": 1,
      "tool": "semantic operation name such as crop, draw_line, draw_box, numeric_calculation, rotation, zoom_in, or read_text",
      "action": "high-level tool action, no large code block",
      "expected_subgoal": "one concise paragraph describing the intermediate evidence this step is meant to obtain",
      "subgoal_result": "detailed factual answer-free result actually obtained, including exact numeric/textual evidence when available and why it matters for this subgoal"
    }
  ],
  "subgoals": ["same ordered strings as tool_steps[*].subgoal_result"]
}

Question:
{question}

Reference final answer, for filtering only. Do not reveal it in the hint:
{ground_truth}

Correct trajectory:
{trajectory}
\end{promptbox}

\section{Additional Experimental Results}

\subsection{Measuring the Tool Effects on Fixed ``Text-easy'' and ``Text-hard'' Sets}
\label{app:fixed_text_easy_hard}
In Section \ref{sec:metrics}, we first split the ``text-easy'' and ``text-hard'' sets based on the behavior of the post-trained agentic visual reasoning model. Therefore, different models may have different splits of the two sets. Although such a design makes the analysis of the adaptiveness achievable, it may make the comparison of tool effects across different models less fair. In this section, we present the tool effects on fixed ``text-easy'' and ``text-hard'' sets, which are split based on the base model (Qwen3-VL-8B-Instruct). Given a dataset, we use the base model to generate five responses for each sample and retain the samples that are answered correctly at least four times as the ``text-easy'' set, and the samples that are answered correctly at most once as the ``text-hard'' set. We analyze models that are post-trained on Qwen3-VL-8B-Instruct: the SFT model trained with \name's SFT data, Metis, and \name. The results are shown in Table \ref{tab:fixed_set}. We report only the tool effects rather than mode adaptiveness, since the difficulty of solving a sample without tools is determined by the base model and may not reflect the capabilities of each post-trained model. Nevertheless, measuring tool gain and harm on these fixed sets remains meaningful, as it reveals how tool-use post-training extends and preserves the capabilities of the base model.

From Table \ref{tab:fixed_set}, we obtain similar observations to those in Section \ref{sec:analysis}, i.e., \name exhibits the largest gap between the tool-induced performance gain and degradation.


\begin{table}[H]
\centering
\caption{\textbf{Detailed results for \TE (\%) on the fixed ``text-easy'' and ``text-hard'' sets.} The definition of all metrics can be found in Section \ref{sec:metrics}. Average denotes the unweighted mean over the four datasets shown.}
\label{tab:fixed_set}
\resizebox{\linewidth}{!}{
\begin{tabular}{l|lccc|cccc}
\toprule
\multirow{2}{*}{\textbf{Dataset}}
& \multirow{2}{*}{\textbf{Method}}
& \multicolumn{3}{c|}{\textbf{Accuracy}}
& \multicolumn{4}{c}{\textbf{Tool Effects}} \\
\cmidrule(lr){3-5}
\cmidrule(lr){6-9}
&
& \textbf{\toolacc $\uparrow$}
& \textbf{\textacc $\uparrow$}
& \textbf{$\Delta_{\mathrm{Acc}}$ $\uparrow$}
& \textbf{\toolgain $\uparrow$}
& \textbf{\toolharm $\downarrow$}
& \textbf{$\Delta_{\mathrm{TE}}$ $\uparrow$}
& \textbf{\textretain $\uparrow$} \\

\midrule
\multirow{3}{*}{HRBench4K}
& SFT
& 81.35 \graytext{(71.25)}
& 77.68 \graytext{(0.00)}
& \textcolor{mygreen}{+3.68}
& 6.65 & 4.68 & \textcolor{mygreen}{+1.98}
& 98.38 \\
& Metis
& 84.13 \graytext{(46.88)}
& 83.80 \graytext{(0.00)}
& \textcolor{mygreen}{+0.33}
& 3.25 & 0.08 & \textcolor{mygreen}{+3.18}
& 95.70 \\
& \textbf{Beacon}
& 83.38 \graytext{(70.65)}
& 80.53 \graytext{(0.00)}
& \textcolor{mygreen}{+2.85}
& 7.05 & 4.48 & \textcolor{mygreen}{+2.58}
& 98.83 \\

\midrule
\multirow{3}{*}{BLINK}
& SFT
& 64.17 \graytext{(72.49)}
& 64.80 \graytext{(0.00)}
& \textcolor{myred}{-0.63}
& 7.80 & 10.46 & \textcolor{myred}{-2.66}
& 93.09 \\
& Metis
& 62.09 \graytext{(24.51)}
& 64.67 \graytext{(0.00)}
& \textcolor{myred}{-2.58}
& 2.70 & 2.97 & \textcolor{myred}{-0.26}
& 83.24 \\
& \textbf{Beacon}
& 65.49 \graytext{(79.79)}
& 64.22 \graytext{(0.00)}
& \textcolor{mygreen}{+1.27}
& 12.98 & 10.63 & \textcolor{mygreen}{+2.36}
& 98.72 \\

\midrule
\multirow{3}{*}{BabyVision}
& SFT
& 14.95 \graytext{(97.32)}
& 17.73 \graytext{(0.00)}
& \textcolor{myred}{-2.78}
& 7.58 & 3.56 & \textcolor{mygreen}{+4.02}
& 100.00 \\
& Metis
& 15.52 \graytext{(15.72)}
& 15.93 \graytext{(0.00)}
& \textcolor{myred}{-0.41}
& 0.36 & 0.21 & \textcolor{mygreen}{+0.15}
& 68.32 \\
& \textbf{Beacon}
& 16.70 \graytext{(95.36)}
& 15.36 \graytext{(0.00)}
& \textcolor{mygreen}{+1.34}
& 6.39 & 2.01 & \textcolor{mygreen}{+4.38}
& 85.71 \\

\midrule
\multirow{3}{*}{MathVista}
& SFT
& 75.76 \graytext{(76.86)}
& 78.82 \graytext{(0.00)}
& \textcolor{myred}{-3.06}
& 4.32 & 6.86 & \textcolor{myred}{-2.54}
& 98.23 \\
& Metis
& 77.36 \graytext{(17.34)}
& 78.02 \graytext{(0.00)}
& \textcolor{myred}{-0.66}
& 0.92 & 0.80 & \textcolor{mygreen}{+0.12}
& 94.68 \\
& \textbf{Beacon}
& 77.80 \graytext{(78.52)}
& 77.18 \graytext{(0.00)}
& \textcolor{mygreen}{+0.62}
& 5.00 & 5.72 & \textcolor{myred}{-0.72}
& 97.68 \\

\midrule
\rowcolor{gray!10}
& SFT
& 59.06 & 59.76
& \textcolor{myred}{-0.70}
& 6.59 & 6.39
& \textcolor{mygreen}{+0.20}
& 97.43 \\
\rowcolor{gray!10}
& Metis
& 59.77 & 60.60
& \textcolor{myred}{-0.83}
& 1.81 & 1.01
& \textcolor{mygreen}{+0.80}
& 85.49 \\
\rowcolor{gray!10}
\multirow{-3}{*}{\textbf{Average}}
& \textbf{Beacon}
& 60.84 & 59.32
& \textcolor{mygreen}{\textbf{+1.52}}
& 7.86 & 5.71
& \textcolor{mygreen}{\textbf{+2.15}}
& 95.24 \\

\bottomrule
\end{tabular}
}
\end{table}
\subsection{Detailed Results of the Adaptiveness and Tool Effects}
\label{app:detailed_metrics}
In Table \ref{tab:metrics_full}, we present the complete results of the \MA and \TE metrics of Beacon and several agentic visual reasoning baselines. 

\begin{table}[htbp]
\centering
\caption{
\textbf{The metrics that reflect the \MA and \TE of agentic visual reasoning models (\%).} The definition of all metrics can be found in Section \ref{sec:metrics}. $\uparrow$ and $\downarrow$ indicate that higher and lower values are better, respectively. The parentheses after the accuracies indicate the proportion of responses that actually use tools under the two different prompts ($p_{\text{tool}}$ and $p_{\text{text}}$).
$\Delta_{\mathrm{Acc}}=\toolacc-\textacc$ and $\Delta_{\mathrm{TE}}=\toolgain-\toolharm$. Positive and non-positive deltas are shown in \textcolor{mygreen}{green} and \textcolor{myred}{red}, respectively.
}
\label{tab:metrics_full}

\resizebox{\linewidth}{!}{
\begin{tabular}{l|lccc|ccc|cccc}
\toprule

\multirow{2}{*}{\textbf{Dataset}}
& \multirow{2}{*}{\textbf{Method}}
& \multicolumn{3}{c|}{\textbf{Accuracy}}
& \multicolumn{3}{c|}{\textbf{Mode Adaptiveness}}
& \multicolumn{4}{c}{\textbf{Tool Effects}} \\

\cmidrule(lr){3-5}
\cmidrule(lr){6-8}
\cmidrule(lr){9-12}

&
& \textbf{\toolacc $\uparrow$}
& \textbf{\textacc $\uparrow$}
& \textbf{$\Delta_{\mathrm{Acc}}$ $\uparrow$}
& \textbf{\MAtool $\uparrow$}
& \textbf{\MAtext $\uparrow$}
& \textbf{\MAmean $\uparrow$}
& \textbf{\toolgain $\uparrow$}
& \textbf{\toolharm $\downarrow$}
& \textbf{$\Delta_{\mathrm{TE}}$ $\uparrow$}
& \textbf{\textretain $\uparrow$} \\

\midrule

\multirow{5}{*}{HRBench4K}
& Thyme
& 76.70 \graytext{(12.78)}
& 76.88 \graytext{(0.00)}
& \textcolor{myred}{-0.18}
& 11.52 & 86.81 & 49.17
& 0.20 & 0.38 & \textcolor{myred}{-0.18}
& 96.06 \\
& DeepEyesV2
& 75.33 \graytext{(98.88)}
& 73.90 \graytext{(0.00)}
& \textcolor{mygreen}{+1.43}
& 99.51 & 1.36 & 50.44
& 7.48 & 6.20 & \textcolor{mygreen}{+1.28}
& 100.00 \\
& CodeV
& 74.05 \graytext{(61.08)}
& 74.78 \graytext{(0.23)}
& \textcolor{myred}{-0.73}
& 43.90 & 33.31 & 38.61
& 3.13 & 3.05 & \textcolor{mygreen}{+0.08}
& 83.73 \\
& Metis
& 84.13 \graytext{(46.88)}
& 83.80 \graytext{(0.00)}
& \textcolor{mygreen}{+0.33}
& 41.77 & 51.85 & 46.81
& 0.75 & 0.33 & \textcolor{mygreen}{+0.43}
& 96.81 \\
& \textbf{Beacon (Ours)}
& 83.38 \graytext{(70.65)}
& 80.53 \graytext{(0.00)}
& \textcolor{mygreen}{+2.85}
& 97.24 & 37.18 & 67.21
& 9.98 & 4.48 & \textcolor{mygreen}{+5.50}
& 100.00 \\

\midrule

\multirow{5}{*}{BLINK}
& Thyme
& 57.41 \graytext{(0.68)}
& 55.35 \graytext{(0.00)}
& \textcolor{mygreen}{+2.06}
& 1.00 & 99.50 & 50.25
& 0.16 & 0.00 & \textcolor{mygreen}{+0.16}
& 84.32 \\
& DeepEyesV2
& 53.83 \graytext{(99.80)}
& 55.27 \graytext{(0.00)}
& \textcolor{myred}{-1.43}
& 99.92 & 0.35 & 50.14
& 10.06 & 9.76 & \textcolor{mygreen}{+0.29}
& 87.50 \\
& CodeV
& 58.56 \graytext{(14.66)}
& 55.98 \graytext{(0.00)}
& \textcolor{mygreen}{+2.58}
& 16.10 & 85.89 & 51.00
& 2.29 & 1.90 & \textcolor{mygreen}{+0.39}
& 79.85 \\
& Metis
& 62.09 \graytext{(24.51)}
& 64.67 \graytext{(0.00)}
& \textcolor{myred}{-2.58}
& 23.09 & 74.42 & 48.76
& 1.27 & 2.25 & \textcolor{myred}{-0.98}
& 87.78 \\
& \textbf{Beacon (Ours)}
& 65.49 \graytext{(79.79)}
& 64.22 \graytext{(0.00)}
& \textcolor{mygreen}{+1.27}
& 93.78 & 28.30 & 61.04
& 13.11 & 11.15 & \textcolor{mygreen}{+1.96}
& 95.28 \\

\midrule

\multirow{5}{*}{BabyVision}
& Thyme
& 12.27 \graytext{(0.62)}
& 11.55 \graytext{(0.00)}
& \textcolor{mygreen}{+0.72}
& 0.72 & 100.00 & 50.36
& 0.00 & 0.00 & \textcolor{myred}{0.00}
& 52.00 \\
& DeepEyesV2
& 11.39 \graytext{(99.90)}
& 1.44 \graytext{(0.00)}
& \textcolor{mygreen}{+9.95}
& 99.90 & 0.00 & 49.95
& 10.98 & 0.88 & \textcolor{mygreen}{+10.10}
& -- \\
& CodeV
& 13.76 \graytext{(0.57)}
& 14.74 \graytext{(0.00)}
& \textcolor{myred}{-0.98}
& 0.69 & 100.00 & 50.35
& 0.00 & 0.00 & \textcolor{myred}{0.00}
& 35.51 \\
& Metis
& 15.52 \graytext{(15.72)}
& 15.93 \graytext{(0.00)}
& \textcolor{myred}{-0.41}
& 17.65 & 94.55 & 56.10
& 0.62 & 0.46 & \textcolor{mygreen}{+0.15}
& 54.23 \\
& \textbf{Beacon (Ours)}
& 16.70 \graytext{(95.36)}
& 15.36 \graytext{(0.00)}
& \textcolor{mygreen}{+1.34}
& 95.99 & 10.77 & 53.38
& 6.44 & 2.42 & \textcolor{mygreen}{+4.02}
& 85.71 \\

\midrule

\multirow{5}{*}{MathVista}
& Thyme
& 71.22 \graytext{(0.10)}
& 71.64 \graytext{(0.00)}
& \textcolor{myred}{-0.42}
& 0.08 & 99.88 & 49.98
& 0.00 & 0.00 & \textcolor{myred}{0.00}
& 93.57 \\
& DeepEyesV2
& 72.22 \graytext{(99.88)}
& 73.68 \graytext{(0.00)}
& \textcolor{myred}{-1.46}
& 100.00 & 0.17 & 50.09
& 5.32 & 7.28 & \textcolor{myred}{-1.96}
& 83.33 \\
& CodeV
& 71.70 \graytext{(0.30)}
& 72.08 \graytext{(0.00)}
& \textcolor{myred}{-0.38}
& 0.76 & 99.86 & 50.31
& 0.04 & 0.04 & \textcolor{myred}{0.00}
& 92.01 \\
& Metis
& 77.36 \graytext{(17.34)}
& 78.02 \graytext{(0.00)}
& \textcolor{myred}{-0.66}
& 21.22 & 83.52 & 52.37
& 0.84 & 1.16 & \textcolor{myred}{-0.32}
& 95.44 \\
& \textbf{Beacon (Ours)}
& 77.80 \graytext{(78.52)}
& 77.18 \graytext{(0.00)}
& \textcolor{mygreen}{+0.62}
& 92.34 & 25.84 & 59.09
& 5.98 & 6.36 & \textcolor{myred}{-0.38}
& 98.46 \\

\midrule

\multirow{5}{*}{TIRBench}
& Thyme
& 20.53 \graytext{(10.45)}
& 19.05 \graytext{(0.02)}
& \textcolor{mygreen}{+1.48}
& 8.07 & 78.57 & 43.32
& 0.58 & 0.25 & \textcolor{mygreen}{+0.33}
& 81.21 \\
& DeepEyesV2
& 15.80 \graytext{(99.11)}
& 20.15 \graytext{(0.00)}
& \textcolor{myred}{-4.35}
& 99.22 & 1.25 & 50.24
& 5.43 & 6.45 & \textcolor{myred}{-1.02}
& 63.64 \\
& CodeV
& 18.11 \graytext{(15.60)}
& 17.04 \graytext{(0.00)}
& \textcolor{mygreen}{+1.07}
& 12.76 & 68.64 & 40.70
& 0.82 & 0.95 & \textcolor{myred}{-0.13}
& 65.86 \\
& Metis
& 21.84 \graytext{(54.32)}
& 21.23 \graytext{(0.00)}
& \textcolor{mygreen}{+0.61}
& 59.32 & 59.23 & 59.28
& 2.21 & 1.28 & \textcolor{mygreen}{+0.92}
& 80.66 \\
& \textbf{Beacon (Ours)}
& 24.28 \graytext{(92.74)}
& 20.56 \graytext{(0.00)}
& \textcolor{mygreen}{+3.72}
& 94.40 & 12.44 & 53.42
& 10.96 & 6.37 & \textcolor{mygreen}{+4.59}
& 75.56 \\

\midrule

\rowcolor{gray!10}
& Thyme
& 47.63 & 46.89 & \textcolor{mygreen}{+0.73}
& 4.28 & 92.95 & 48.62
& 0.19 & 0.12 & \textcolor{mygreen}{+0.06}
& 81.43 \\
\rowcolor{gray!10}
& DeepEyesV2
& 45.71 & 44.89 & \textcolor{mygreen}{+0.83}
& 99.71 & 0.63 & 50.17
& 7.85 & 6.11 & \textcolor{mygreen}{+1.74}
& -- \\
\rowcolor{gray!10}
& CodeV
& 47.24 & 46.92 & \textcolor{mygreen}{+0.31}
& 14.84 & 77.54 & 46.19
& 1.26 & 1.19 & \textcolor{mygreen}{+0.07}
& 71.39 \\
\rowcolor{gray!10}
& Metis
& 52.19 & 52.73 & \textcolor{myred}{-0.54}
& 32.61 & 72.71 & 52.66
& 1.14 & 1.10 & \textcolor{mygreen}{+0.04}
& 82.98 \\
\rowcolor{gray!10}
\multirow{-5}{*}{\textbf{Average}}
& \textbf{Beacon (Ours)}
& 53.53 & 51.57 & \textcolor{mygreen}{\textbf{+1.96}}
& 94.75 & 22.91 & \textbf{58.83}
& 9.29 & 6.16 & \textcolor{mygreen}{\textbf{+3.14}}
& \textbf{91.00} \\

\bottomrule
\end{tabular}
}
\end{table}

\subsection{Detailed Results of Ablation Study}
\label{app:ablation}
We provide detailed results of the ablation study in Table~\ref{tab:detailed_ablation}.


\begin{table}[H]
\centering
\caption{\textbf{Detailed ablation results for \MA and \TE (\%).} The definition of all metrics can be found in Section \ref{sec:metrics}.}
\label{tab:detailed_ablation}

\resizebox{\linewidth}{!}{
\begin{tabular}{l|lccc|ccc|cccc}
\toprule

\multirow{2}{*}{\textbf{Dataset}}
& \multirow{2}{*}{\textbf{Method}}
& \multicolumn{3}{c|}{\textbf{Accuracy}}
& \multicolumn{3}{c|}{\textbf{Mode Adaptiveness}}
& \multicolumn{4}{c}{\textbf{Tool Effects}} \\

\cmidrule(lr){3-5}
\cmidrule(lr){6-8}
\cmidrule(lr){9-12}

&
& \textbf{\toolacc $\uparrow$}
& \textbf{\textacc $\uparrow$}
& \textbf{$\Delta_{\mathrm{Acc}}$ $\uparrow$}
& \textbf{\MAtool $\uparrow$}
& \textbf{\MAtext $\uparrow$}
& \textbf{\MAmean $\uparrow$}
& \textbf{\toolgain $\uparrow$}
& \textbf{\toolharm $\downarrow$}
& \textbf{$\Delta_{\mathrm{TE}}$ $\uparrow$}
& \textbf{\textretain $\uparrow$} \\

\midrule

\multirow{5}{*}{HRBench4K}
& SFT
& 81.35 \graytext{(71.25)}
& 77.68 \graytext{(0.00)}
& \textcolor{mygreen}{+3.68}
& 95.04 & 33.57 & 64.31
& 6.33 & 3.53 & \textcolor{mygreen}{+2.80}
& 98.91 \\

& GRPO
& 82.55 \graytext{(70.85)}
& 80.10 \graytext{(0.00)}
& \textcolor{mygreen}{+2.45}
& 94.29 & 33.94 & 64.12
& 9.23 & 4.48 & \textcolor{mygreen}{+4.75}
& 99.82 \\

& GRPO+NAAR
& 82.70 \graytext{(70.33)}
& 80.50 \graytext{(0.00)}
& \textcolor{mygreen}{+2.20}
& 97.72 & 38.00 & 67.86
& 9.15 & 4.53 & \textcolor{mygreen}{+4.63}
& 99.22 \\

& GRPO+HCE
& 83.20 \graytext{(70.88)}
& 80.00 \graytext{(0.00)}
& \textcolor{mygreen}{+3.20}
& 98.05 & 37.91 & 67.98
& 9.65 & 4.00 & \textcolor{mygreen}{+5.65}
& 99.56 \\

& \textbf{Beacon}
& 83.38 \graytext{(70.65)}
& 80.53 \graytext{(0.00)}
& \textcolor{mygreen}{+2.85}
& 97.24 & 37.18 & 67.21
& 9.98 & 4.48 & \textcolor{mygreen}{+5.50}
& 100.00 \\

\midrule

\multirow{5}{*}{BLINK}
& SFT
& 64.17 \graytext{(72.49)}
& 64.80 \graytext{(0.00)}
& \textcolor{myred}{-0.63}
& 86.17 & 27.33 & 56.75
& 6.61 & 7.48 & \textcolor{myred}{-0.87}
& 95.01 \\

& GRPO
& 65.48 \graytext{(79.55)}
& 63.91 \graytext{(0.00)}
& \textcolor{mygreen}{+1.57}
& 92.79 & 23.78 & 58.29
& 11.06 & 9.36 & \textcolor{mygreen}{+1.69}
& 95.70 \\

& GRPO+NAAR
& 65.70 \graytext{(79.66)}
& 64.06 \graytext{(0.00)}
& \textcolor{mygreen}{+1.64}
& 93.85 & 29.13 & 61.49
& 11.36 & 10.11 & \textcolor{mygreen}{+1.25}
& 95.37 \\

& GRPO+HCE
& 65.24 \graytext{(79.48)}
& 64.18 \graytext{(0.00)}
& \textcolor{mygreen}{+1.06}
& 92.70 & 27.31 & 60.01
& 11.29 & 9.21 & \textcolor{mygreen}{+2.08}
& 96.05 \\

& \textbf{Beacon}
& 65.49 \graytext{(79.79)}
& 64.22 \graytext{(0.00)}
& \textcolor{mygreen}{+1.27}
& 93.78 & 28.30 & 61.04
& 13.11 & 11.15 & \textcolor{mygreen}{+1.96}
& 95.28 \\

\midrule

\multirow{5}{*}{BabyVision}
& SFT
& 14.95 \graytext{(97.32)}
& 17.73 \graytext{(0.00)}
& \textcolor{myred}{-2.78}
& 97.31 & 8.57 & 52.94
& 6.13 & 3.92 & \textcolor{mygreen}{+2.22}
& 75.00 \\

& GRPO
& 15.10 \graytext{(94.69)}
& 14.59 \graytext{(0.00)}
& \textcolor{mygreen}{+0.52}
& 92.46 & 8.36 & 50.41
& 5.93 & 5.36 & \textcolor{mygreen}{+0.57}
& 68.00 \\

& GRPO+NAAR
& 16.96 \graytext{(95.10)}
& 16.60 \graytext{(0.00)}
& \textcolor{mygreen}{+0.36}
& 95.79 & 14.71 & 55.25
& 5.72 & 3.97 & \textcolor{mygreen}{+1.75}
& 80.00 \\

& GRPO+HCE
& 16.91 \graytext{(94.74)}
& 15.21 \graytext{(0.00)}
& \textcolor{mygreen}{+1.70}
& 95.23 & 9.49 & 52.36
& 7.58 & 4.54 & \textcolor{mygreen}{+3.04}
& 82.01 \\

& \textbf{Beacon}
& 16.70 \graytext{(95.36)}
& 15.36 \graytext{(0.00)}
& \textcolor{mygreen}{+1.34}
& 95.99 & 10.77 & 53.38
& 6.44 & 2.42 & \textcolor{mygreen}{+4.02}
& 85.71 \\

\midrule

\multirow{5}{*}{MathVista}
& SFT
& 75.76 \graytext{(76.86)}
& 78.82 \graytext{(0.00)}
& \textcolor{myred}{-3.06}
& 90.43 & 22.81 & 56.62
& 2.90 & 6.38 & \textcolor{myred}{-3.48}
& 97.98 \\

& GRPO
& 76.60 \graytext{(78.76)}
& 77.50 \graytext{(0.00)}
& \textcolor{myred}{-0.90}
& 90.64 & 23.59 & 57.12
& 5.26 & 7.04 & \textcolor{myred}{-1.78}
& 98.23 \\

& GRPO+NAAR
& 77.22 \graytext{(78.50)}
& 77.68 \graytext{(0.00)}
& \textcolor{myred}{-0.46}
& 93.95 & 26.49 & 60.22
& 5.64 & 6.44 & \textcolor{myred}{-0.80}
& 98.39 \\

& GRPO+HCE
& 76.30 \graytext{(78.20)}
& 77.28 \graytext{(0.00)}
& \textcolor{myred}{-0.98}
& 92.61 & 26.97 & 59.79
& 5.50 & 7.00 & \textcolor{myred}{-1.50}
& 98.32 \\

& \textbf{Beacon}
& 77.80 \graytext{(78.52)}
& 77.18 \graytext{(0.00)}
& \textcolor{mygreen}{+0.62}
& 92.34 & 25.84 & 59.09
& 5.98 & 6.36 & \textcolor{myred}{-0.38}
& 98.46 \\

\midrule

\multirow{5}{*}{TIRBench}
& SFT
& 23.85 \graytext{(90.83)}
& 20.21 \graytext{(0.03)}
& \textcolor{mygreen}{+3.64}
& 93.29 & 7.46 & 50.38
& 10.30 & 2.93 & \textcolor{mygreen}{+7.37}
& 83.06 \\

& GRPO
& 21.53 \graytext{(92.12)}
& 20.89 \graytext{(0.00)}
& \textcolor{mygreen}{+0.64}
& 93.15 & 10.01 & 51.58
& 9.09 & 7.36 & \textcolor{mygreen}{+1.73}
& 70.03 \\

& GRPO+NAAR
& 23.65 \graytext{(93.19)}
& 22.35 \graytext{(0.00)}
& \textcolor{mygreen}{+1.30}
& 95.18 & 12.02 & 53.60
& 12.15 & 6.26 & \textcolor{mygreen}{+5.89}
& 71.76 \\

& GRPO+HCE
& 23.05 \graytext{(91.80)}
& 22.44 \graytext{(0.00)}
& \textcolor{mygreen}{+0.61}
& 93.21 & 10.14 & 51.68
& 11.85 & 6.34 & \textcolor{mygreen}{+5.51}
& 72.69 \\

& \textbf{Beacon}
& 24.28 \graytext{(92.74)}
& 20.56 \graytext{(0.00)}
& \textcolor{mygreen}{+3.72}
& 94.40 & 12.44 & 53.42
& 10.96 & 6.37 & \textcolor{mygreen}{+4.59}
& 75.56 \\

\midrule

\rowcolor{gray!10}
& SFT
& 52.02 & 51.85
& \textcolor{mygreen}{+0.17}
& 92.45 & 19.95 & 56.20
& 6.45 & 4.85
& \textcolor{mygreen}{+1.61}
& 89.99 \\

\rowcolor{gray!10}
& GRPO
& 52.25 & 51.40
& \textcolor{mygreen}{+0.86}
& 92.67 & 19.94 & 56.30
& 8.11 & 6.72
& \textcolor{mygreen}{+1.39}
& 86.36 \\

\rowcolor{gray!10}
& GRPO+NAAR
& 53.25 & 52.24
& \textcolor{mygreen}{+1.01}
& 95.30 & 24.07 & \textbf{59.68}
& 8.80 & 6.26
& \textcolor{mygreen}{+2.54}
& 88.95 \\

\rowcolor{gray!10}
& GRPO+HCE
& 52.94 & 51.82
& \textcolor{mygreen}{+1.12}
& 94.36 & 22.36 & 58.36
& 9.17 & 6.22
& \textcolor{mygreen}{+2.96}
& 89.73 \\

\rowcolor{gray!10}
\multirow{-5}{*}{\textbf{Average}}
& \textbf{Beacon}
& 53.53 & 51.57
& \textcolor{mygreen}{\textbf{+1.96}}
& 94.75 & 22.91 & 58.83
& 9.29 & 6.16
& \textcolor{mygreen}{\textbf{+3.14}}
& \textbf{91.00} \\

\bottomrule
\end{tabular}
}
\end{table}

\subsection{Training Dynamics of the RL Stage}
\label{app:rl_dynamics}
In this section, we provide a comprehensive analysis of the training dynamics of the RL process.

\begin{figure*}[t]
    \centering
    \begin{subfigure}[t]{0.32\textwidth}
        \centering
        \includegraphics[width=\linewidth]{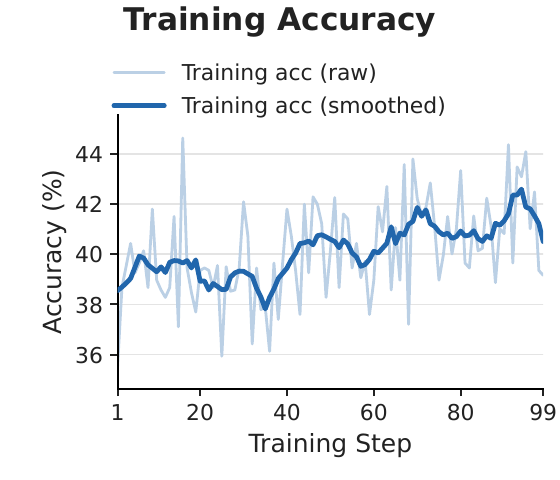}
        \caption{}
        \label{fig:training_accuracy}
    \end{subfigure}\hfill
    \begin{subfigure}[t]{0.32\textwidth}
        \centering
        \includegraphics[width=\linewidth]{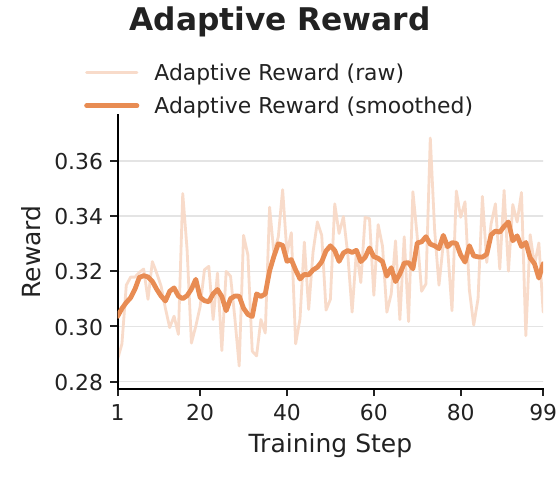}
        \caption{}
        \label{fig:adaptive_training_reward}
    \end{subfigure}\hfill
    \begin{subfigure}[t]{0.32\textwidth}
        \centering
        \includegraphics[width=\linewidth]{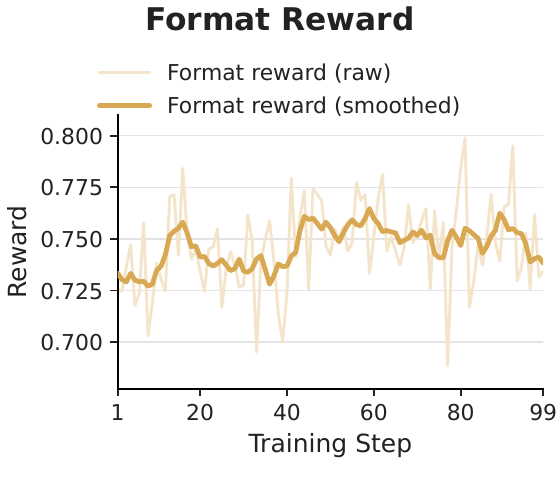}
        \caption{}
        \label{fig:format_reward}
    \end{subfigure}

    \caption{\textbf{Training accuracy and rewards of the RL stage.} \textbf{(a)} Mean training accuracy. \textbf{(b)} Adaptive reward. \textbf{(c)} Format reward.}
    \label{fig:training_rewards}
\end{figure*}

\begin{figure*}[t]
    \centering

    \begin{subfigure}[t]{0.32\textwidth}
        \centering
        \includegraphics[width=\linewidth]{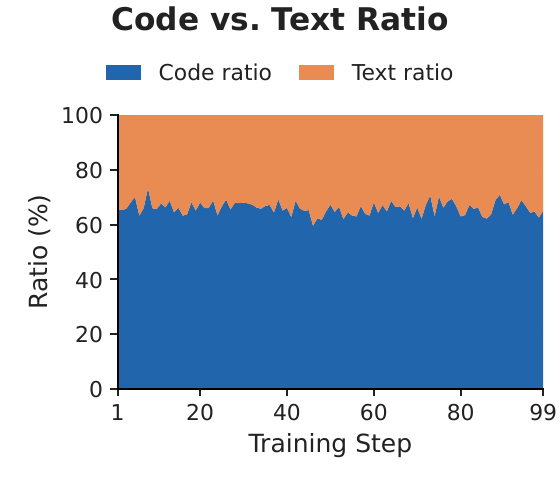}
        \caption{}
        \label{fig:code_text_response_ratio}
    \end{subfigure}\hfill
    \begin{subfigure}[t]{0.32\textwidth}
        \centering
        \includegraphics[width=\linewidth]{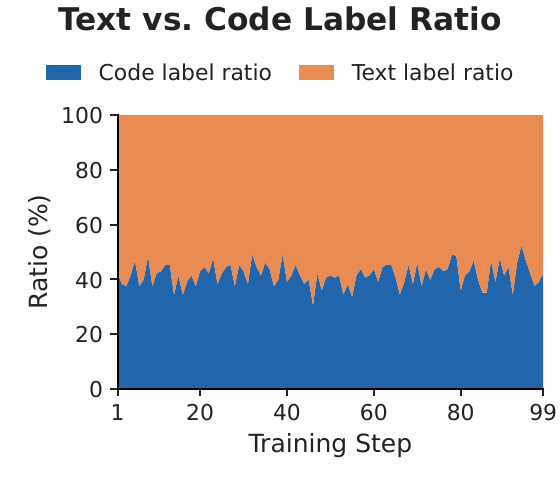}
        \caption{}
        \label{fig:text_code_label_ratio}
    \end{subfigure}
    \begin{subfigure}[t]{0.32\textwidth}
        \centering
        \includegraphics[width=\linewidth]{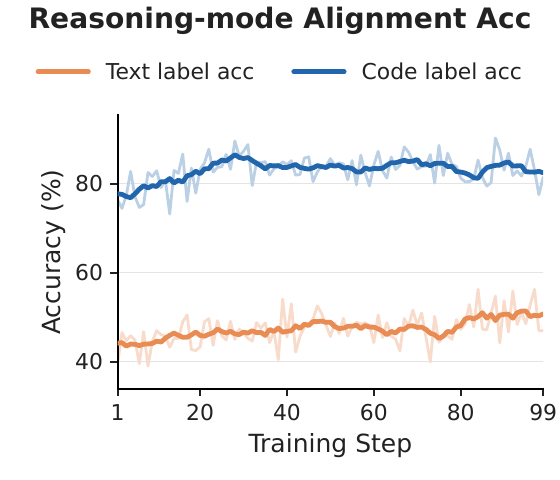}
        \caption{}
        \label{fig:alignment}
    \end{subfigure}

    \caption{\textbf{(a)} Proportions of pure-text and code-assisted responses during training. \textbf{(b)} Proportions of training groups assigned the ``text'' and ``code'' adaptive labels for reward computation. \textbf{(c)} Accuracy of reasoning-mode selection, measured as the proportion of responses whose reasoning mode matches the adaptive label of their corresponding group.}
    \label{fig:code_text_ratio}
\end{figure*}

\paragraph{Training rewards and adaptive behavior.}
We present the reward dynamics during the RL stage in Figure~\ref{fig:training_rewards}, where the training rewards generally improve during training. Figure~\ref{fig:code_text_response_ratio} shows the proportions of pure-text and code-assisted responses, which remain relatively stable throughout training, with no clear shift toward either reasoning mode. Figure \ref{fig:text_code_label_ratio} further shows that the ratio between the adaptive labels of both reasoning modes remains steady during training. In contrast, Figure~\ref{fig:alignment} reports the reasoning-mode alignment accuracy, defined as the proportion of rollout trajectories whose actual reasoning mode matches the adaptive label assigned to their group. This accuracy consistently improves over the course of training, suggesting that the model gradually learns an adaptive tool-invocation policy rather than simply developing a preference for one reasoning mode.

\begin{figure*}[t]
    \centering

    \begin{subfigure}[t]{0.48\textwidth}
        \centering
        \includegraphics[width=\linewidth]{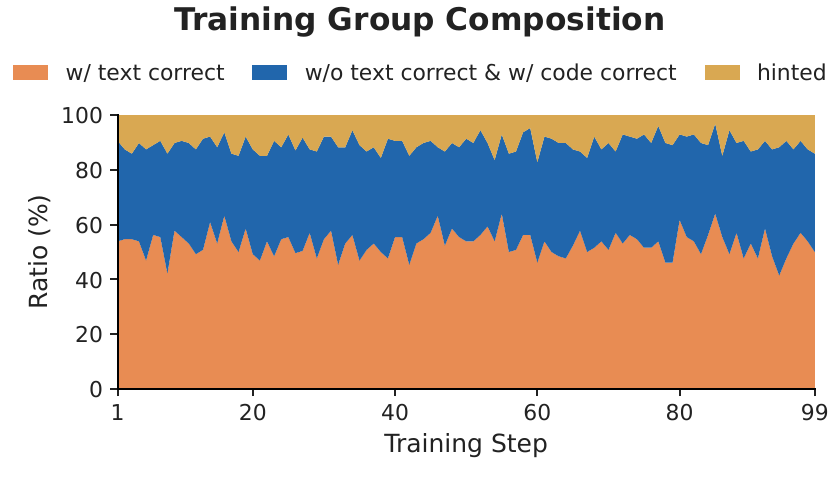}
        \caption{Composition of training groups.}
        \label{fig:group_composition}
    \end{subfigure}\hfill
    \begin{subfigure}[t]{0.48\textwidth}
        \centering
        \includegraphics[width=\linewidth]{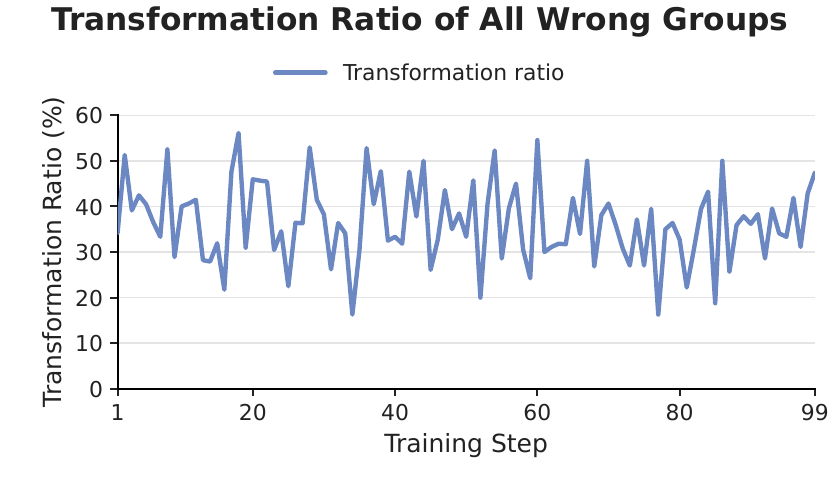}
        \caption{Transformation.}
        \label{fig:transformation_ratio}
    \end{subfigure}

    \caption{\textbf{(a)} Proportions of groups containing at least one correct pure-text response (``w/ text correct''), groups containing no correct pure-text response but at least one correct code-assisted response (``w/o text correct \& w/ code correct''), and hint-induced groups (``hinted''). \textbf{(b)} Proportion of initially all-incorrect groups that are converted into groups with non-zero accuracy advantages through hint-guided re-rollouts.}
    \label{fig:group_and_transform_ratio}
\end{figure*}

\paragraph{Training Group Composition and All-Wrong Group Transformation.}
Figure~\ref{fig:group_composition} presents the composition of training rollout groups during the RL stage. We categorize the groups into three types: groups containing at least one correct pure-text response (``w/ text correct''), groups containing no correct pure-text response but at least one correct code-assisted response (``w/o text correct \& w/ code correct''), and hint-induced groups whose responses are initially all incorrect (``hinted''). These three categories account for relatively stable proportions of approximately 50\%, 35\%, and 15\%, respectively. Figure~\ref{fig:transformation_ratio} further reports the proportion of initially all-incorrect groups that are converted into groups with non-zero accuracy advantages through hint-guided re-rollouts. Approximately 40\% of such groups can be successfully recycled by our HCE mechanism, demonstrating its effectiveness in recovering useful learning signals from otherwise uninformative rollout groups.

\section{Discussion}
\subsection{Limitations}
\label{app:limitations}
Our diagnostics have several limitations. First, the ``text-easy'' and ``text-hard'' sets are defined from a finite number of stochastic rollouts and are model-dependent; we mitigate this issue with five rollouts and additionally report results on a fixed base-model split, but the categories remain an empirical proxy for difficulty. Second, our implementation studies one tool interface—a sandboxed Python environment—so the findings may not transfer directly to web, retrieval, or embodied tools. Third, the HCE module relies on Gemini 3.1 Pro to generate training-time hints, which could be potentially costly. Extending the evaluation to causal tool-effect estimates, broader tool ecosystems, and explicit efficiency measurements is important future work.

\subsection{Failed Attempts}

We also explored a more aggressive adaptive reward scheme. 
In addition to the standard rollout group, we force the model to generate a group of pure-text reasoning trajectories. 
If either the standard rollout group or the forced-text rollout group contains at least one correct pure-text trajectory, the adaptive label is assigned as ``text.'' 
However, this strategy strongly biases the policy toward pure-text reasoning, causing the code-use ratio to collapse and substantially degrading performance on perception-intensive benchmarks such as V$^{*}$ and HRBench.

The root cause is that the additional forced-text rollouts artificially increase the probability of observing at least one successful pure-text trajectory, even when pure-text reasoning is not the most reliable strategy on average. 
Consequently, the group-level label reflects the best outcome across multiple attempts rather than the reasoning mode most likely to succeed in a single rollout. 
This introduces a mismatch between training and evaluation: a problem may produce one correct pure-text trajectory among multiple training rollouts, yet still have a low probability of being solved by a single pure-text rollout at test time. 
As a result, the model becomes overly biased toward pure-text reasoning, leading to inferior test-time performance.

\section{Case Study}
\subsection{Reasoning Trajectory Examples}

\begin{table*}[t]
\centering
\small

\begin{tabular}{
@{}p{0.38\textwidth}
@{\hspace{0.03\textwidth}}
p{0.58\textwidth}@{}
}

\begin{minipage}[t]{\linewidth}
\vspace{0pt}
\centering

\includegraphics[
    width=\linewidth,
    height=5cm,
    keepaspectratio
]{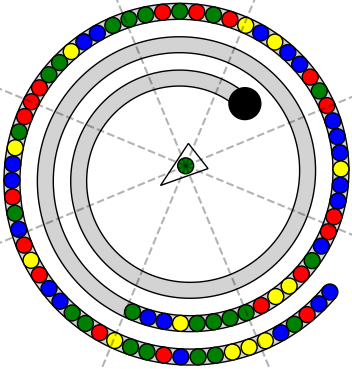}\\[-0.2em]
{\scriptsize Original Image (q0)}

\vspace{0.5em}

\includegraphics[
    width=\linewidth,
    height=5cm,
    keepaspectratio
]{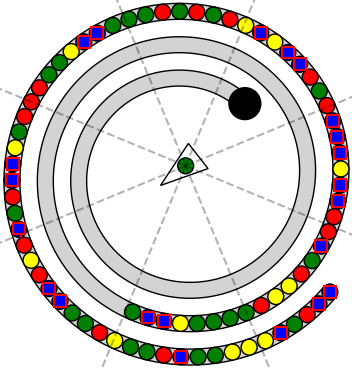}\\[-0.2em]
{\scriptsize g0}

\vspace{0.5em}

\includegraphics[
    width=\linewidth,
    height=5cm,
    keepaspectratio
]{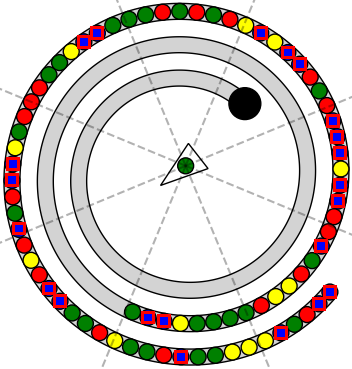}\\[-0.2em]
{\scriptsize g1}

\end{minipage}
&
\begin{minipage}[t]{\linewidth}
\vspace{0pt}

\textbf{Question}\\
How many blue marbles are there on the track?

\vspace{0.6em}

\textbf{Beacon Reasoning}\\
\texttt{<tool\_call>}\\
\texttt{img = get\_image("q0")}\\
\texttt{img.size}\\
\texttt{</tool\_call>}\\
\texttt{<tool\_response>}\\
\texttt{[352, 369]}\\
\texttt{Available image refs: q0}\\
\texttt{</tool\_response>}\\
\texttt{<observation>}\\
\texttt{The image size is 352x369. I will now write a script to detect the blue marbles by finding connected components of blue pixels and filtering them based on size and color.}\\
\texttt{</observation>}\\
\texttt{<tool\_call>}\\
\texttt{import numpy as np}\\
\texttt{from PIL import Image}\\
\texttt{img = get\_image("q0")}\\
\texttt{img\_np = np.array(img)}\\
\texttt{\# Create a mask for blue pixels}\\
\texttt{blue\_mask = (img\_np[:, :, 0] < 100) \&}\\
\hspace*{2em}\texttt{(img\_np[:, :, 1] < 150) \&}\\
\hspace*{2em}\texttt{(img\_np[:, :, 2] > 200)}\\
\texttt{\# Simple connected components using BFS}\\
\texttt{def count\_components(mask):}\\
\hspace*{2em}\texttt{visited = np.zeros\_like(mask, dtype=np.bool\_)}\\
\hspace*{2em}\texttt{count = 0}\\
\hspace*{2em}\texttt{h, w = mask.shape}\\
\hspace*{2em}\texttt{for i in range(h):}\\
\hspace*{4em}\texttt{for j in range(w):}\\
\hspace*{6em}\texttt{if mask[i, j] and not visited[i, j]:}\\
\hspace*{8em}\texttt{count += 1}\\
\hspace*{8em}\texttt{queue = [(i, j)]}\\
\hspace*{8em}\texttt{visited[i, j] = True}\\
\texttt{\ldots}\\
\texttt{Available image refs: q0, g0, g1}\\
\texttt{</tool\_response>}\\
\texttt{<observation>}\\
\texttt{The bounding boxes for the 22 valid components perfectly enclose all the blue marbles on the track. There are no missed marbles and no false positives. The count is exactly 22.}\\
\texttt{</observation>}\\
\texttt{<answer>}\\
\texttt{22}\\
\texttt{</answer>}

\vspace{0.6em}

\textbf{Beacon Answer}\\
\textcolor{blue}{22 (\checkmark)}

\end{minipage}

\end{tabular}

\caption{\textbf{Reasoning trajectory example 1: Pixel-level computation.}}
\label{tab:case_1}
\end{table*}
\begin{table*}[t]
\centering
\small

\begin{tabular}{
@{}p{0.38\textwidth}
@{\hspace{0.03\textwidth}}
p{0.58\textwidth}@{}
}

\begin{minipage}[t]{\linewidth}
\vspace{0pt}
\centering

\includegraphics[
    width=\linewidth,
    height=3.2cm,
    keepaspectratio
]{figure/case/1/1.png}\\[-0.2em]
{\scriptsize Original Image (q0)}

\vspace{0.5em}

\includegraphics[
    width=\linewidth,
    height=3.2cm,
    keepaspectratio
]{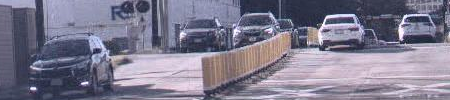}\\[-0.2em]
{\scriptsize g0}

\vspace{0.5em}

\includegraphics[
    width=\linewidth,
    height=3.2cm,
    keepaspectratio
]{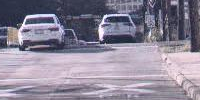}\\[-0.2em]
{\scriptsize g1}

\vspace{0.5em}

\includegraphics[
    width=\linewidth,
    height=3.2cm,
    keepaspectratio
]{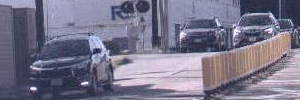}\\[-0.2em]
{\scriptsize g2}

\vspace{0.5em}

\includegraphics[
    width=\linewidth,
    height=3.2cm,
    keepaspectratio
]{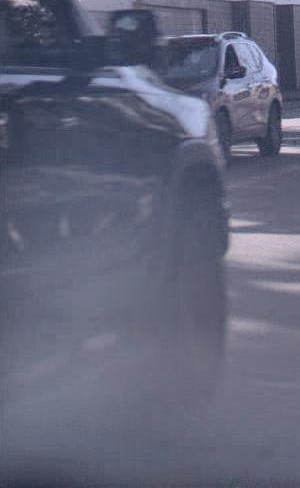}\\[-0.2em]
{\scriptsize g3}

\vspace{0.5em}

\includegraphics[
    width=\linewidth,
    height=3.2cm,
    keepaspectratio
]{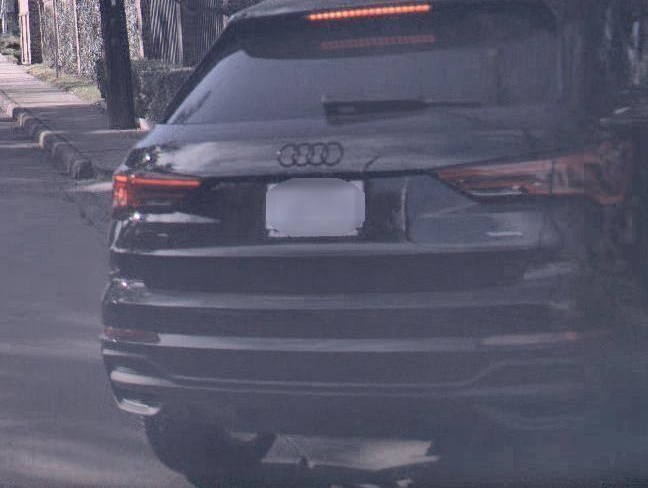}\\[-0.2em]
{\scriptsize g4}

\end{minipage}
&
\begin{minipage}[t]{\linewidth}
\vspace{0pt}

\textbf{Question}\\
How many cars are white in this picture?

\vspace{0.6em}

\textbf{Beacon Reasoning}\\
\texttt{<tool\_call>}\\
\texttt{img = get\_image("q0")}\\
\texttt{img.size}\\
\texttt{</tool\_call>}\\
\texttt{<tool\_response>}\\
\texttt{[1448, 938]}\\
\texttt{Available image refs: q0}\\
\texttt{</tool\_response>}\\
\texttt{<observation>}\\
\texttt{The image size is 1448x938. I will crop the middle section to get a better look at the cars in the distance.}\\
\texttt{</observation>}\\
\texttt{<tool\_call>}\\
\texttt{img = get\_image("q0")}\\
\texttt{cropped = img.crop((300, 450, 750, 550))}\\
\texttt{cropped}\\
\texttt{</tool\_call>}\\
\texttt{<tool\_response>}\\
\texttt{Returned image ref: g0}\\
\texttt{Available image refs: q0, g0}\\
\texttt{</tool\_response>}\\
\texttt{<observation>}\\
\texttt{In the cropped image, I can see several cars. Let's look closer at the cars in the distance to determine their colors.}\\
\texttt{</observation>}\\
\texttt{<tool\_call>}\\
\texttt{img = get\_image("q0")}\\
\texttt{cropped = img.crop((600, 450, 800, 550))}\\
\texttt{cropped}\\
\texttt{</tool\_call>}\\
\texttt{<tool\_response>}\\
\texttt{Returned image ref: g1}\\
\texttt{Available image refs: q0, g0, g1}\\
\texttt{\ldots}\\
\texttt{Returned image ref: g4}\\
\texttt{Available image refs: q0, g0, g1, g2, g3, g4}\\
\texttt{</tool\_response>}\\
\texttt{<observation>}\\
\texttt{The car on the right side of the foreground is dark grey or black. So, the only white cars are the two parked on the right side of the street in the distance. Therefore, there are 2 white cars in the picture.}\\
\texttt{</observation>}\\
\texttt{<answer>}\\
\texttt{B}\\
\texttt{</answer>}

\vspace{0.6em}

\textbf{Beacon Answer}\\
\textcolor{blue}{B (\checkmark)}

\end{minipage}

\end{tabular}

\caption{\textbf{Reasoning trajectory example 2: Visual search and counting.}}
\label{tab:case_2}
\end{table*}
\begin{table*}[t]
\centering
\small

\begin{tabular}{
@{}p{0.38\textwidth}
@{\hspace{0.03\textwidth}}
p{0.58\textwidth}@{}
}

\begin{minipage}[t]{\linewidth}
\vspace{0pt}
\centering

\includegraphics[
    width=\linewidth,
    height=3.2cm,
    keepaspectratio
]{figure/case/3/1.jpeg}\\[-0.2em]
{\scriptsize Original Image (q0)}

\vspace{0.5em}

\includegraphics[
    width=\linewidth,
    height=3.2cm,
    keepaspectratio
]{figure/case/3/2.jpeg}\\[-0.2em]
{\scriptsize Original Image (q1)}

\vspace{0.5em}

\includegraphics[
    width=\linewidth,
    height=3.2cm,
    keepaspectratio
]{figure/case/3/3.jpeg}\\[-0.2em]
{\scriptsize Original Image (q2)}

\vspace{0.5em}

\includegraphics[
    width=\linewidth,
    height=3.2cm,
    keepaspectratio
]{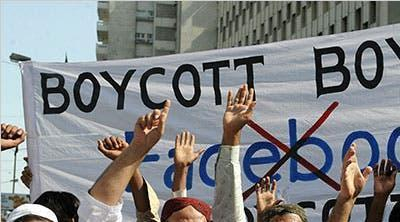}\\[-0.2em]
{\scriptsize g0}

\end{minipage}
&
\begin{minipage}[t]{\linewidth}
\vspace{0pt}

\textbf{Question}\\
Given the first image with the lower right corner missing, can you tell which one of the second image or the third image is the missing part?

\vspace{0.6em}

\textbf{Beacon Reasoning}\\
\texttt{<tool\_call>}\\
\texttt{import numpy as np}\\
\texttt{from PIL import Image}\\
\texttt{img1 = get\_image("q0")}\\
\texttt{img2 = get\_image("q1")}\\
\texttt{img3 = get\_image("q2")}\\
\texttt{print(f"Image 1 size: \{img1.size\}")}\\
\texttt{print(f"Image 2 size: \{img2.size\}")}\\
\texttt{print(f"Image 3 size: \{img3.size\}")}\\
\texttt{\# Let's crop the bottom-right corner of Image 1 and compare}\\
\texttt{\# it with Image 2 and Image 3.}\\
\texttt{\# The missing part is a black rectangle in the lower-right}\\
\texttt{\# corner. Let's find its coordinates.}\\
\texttt{</tool\_call>}\\
\texttt{<tool\_response>}\\
\texttt{Image 1 size: (400, 222)}\\
\texttt{Image 2 size: (200, 111)}\\
\texttt{Image 3 size: (200, 111)}\\
\texttt{Available image refs: q0, q1, q2}\\
\texttt{</tool\_response>}\\
\texttt{<observation>}\\
\texttt{The first image has a black rectangle in the lower-right corner (from x=200 to 400, y=111 to 222). The second and third images are exactly half the size of the first image. I will extract the bottom-right quadrant of the first image and compare it with the second and third images.}\\
\texttt{…}\\
\texttt{<observation>}\\
\texttt{Pasting Image 3 into the missing spot of Image 1 creates a seamless and logical continuation of the protest scene. The banner with "cebo" crossed out perfectly aligns with the text seen in Image 1, and the hands of the protesters also align well. The second image, showing two men, would not fit this context at all. Therefore, Image 3 is the correct missing part.}\\
\texttt{</observation>}\\
\texttt{<answer>}\\
\texttt{B}\\
\texttt{</answer>}

\vspace{0.6em}

\textbf{Beacon Answer}\\
\textcolor{blue}{B (\checkmark)}

\end{minipage}

\end{tabular}

\caption{
\textbf{Reasoning trajectory example 3: Jigsaw.}
}
\label{tab:case_3}
\end{table*}

As illustrated in Tables~\ref{tab:case_1}, \ref{tab:case_2}, and \ref{tab:case_3}, Beacon effectively leverages a diverse set of visual tools to solve complex reasoning tasks.

\subsection{Hint Examples}

\begin{table*}[t]
\centering
\small

\begin{tabular}{
@{}p{0.38\textwidth}
@{\hspace{0.03\textwidth}}
p{0.58\textwidth}@{}
}

\begin{minipage}[t]{\linewidth}
\vspace{0pt}
\centering

\includegraphics[
    width=\linewidth,
    height=3.2cm,
    keepaspectratio
]{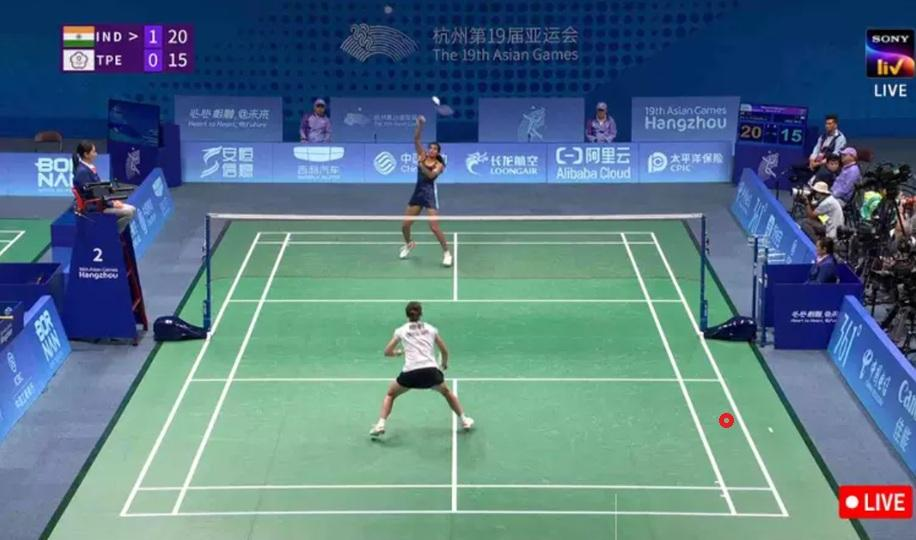}\\[-0.2em]
{\scriptsize Original Image (q0)}

\vspace{0.5em}

\includegraphics[
    width=\linewidth,
    height=3.2cm,
    keepaspectratio
]{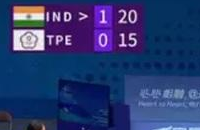}\\[-0.2em]
{\scriptsize g0}

\vspace{0.5em}

\includegraphics[
    width=\linewidth,
    height=3.2cm,
    keepaspectratio
]{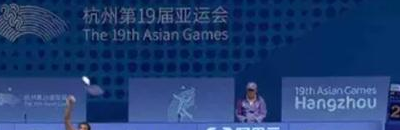}\\[-0.2em]
{\scriptsize g1}

\vspace{0.5em}

\includegraphics[
    width=\linewidth,
    height=3.2cm,
    keepaspectratio
]{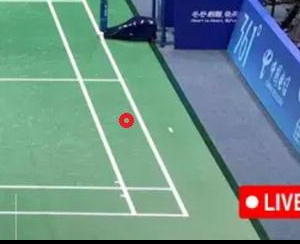}\\[-0.2em]
{\scriptsize g2}

\vspace{0.5em}

\includegraphics[
    width=\linewidth,
    height=3.2cm,
    keepaspectratio
]{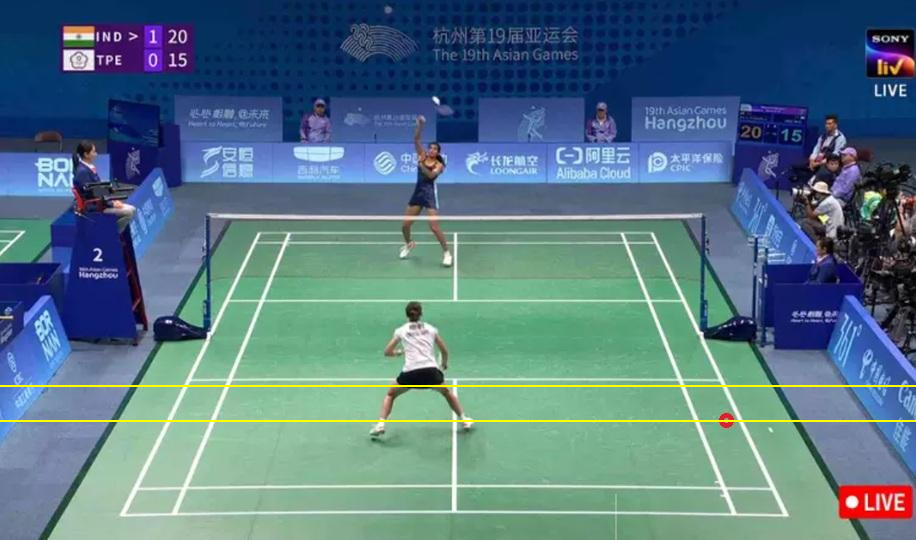}\\[-0.2em]
{\scriptsize g3}

\end{minipage}
&
\begin{minipage}[t]{\linewidth}
\vspace{0pt}

\textbf{Question}\\
In which year was this event held? The player in the blue jersey is from India, and she is executing a badminton smash. If the shuttlecock lands inside the red circle, what will the new match score be?

\vspace{0.6em}

\textbf{Hint}\\
1. [Instruction] Crop the scoreboard and red circle areas from the image; [Expected Subgoal] Determine the current score, event name, and the boundary status of the red circle;\\
2. [Instruction] Calculate the new score based on the out-of-bounds rule and identify the year of the event; [Expected Subgoal] Find the updated match score and the year the event took place;

\vspace{0.6em}

\textbf{Beacon Reasoning Given the Hints}\\
\texttt{<tool\_call>}\\
\texttt{img = get\_image("q0")}\\
\texttt{\ldots}\\
\texttt{<tool\_call>}\\
\texttt{from PIL import ImageDraw}\\
\texttt{annotated = get\_image("q0").copy()}\\
\texttt{draw = ImageDraw.Draw(annotated)}\\
\texttt{draw.line((0, 385, 928, 385), fill="yellow", width=2)}\\
\texttt{draw.line((0, 420, 928, 420), fill="yellow", width=2)}\\
\texttt{annotated}\\
\texttt{</tool\_call>}\\
\texttt{<tool\_response>}\\
\texttt{generated\_images: g3}\\
\texttt{Original image refs: q0}\\
\texttt{Generated image refs: g0, g1, g2, g3}\\
\texttt{</tool\_response>}\\
\texttt{<observation>}\\
\texttt{The red circle is located at the bottom-right corner of the court, which is outside the singles court boundary and in the alley. A smash that lands in the alley is considered out of bounds, so it is a fault for the player who hit it. The current score is IND 20, TPE 0 15. This means IND has won Game 1 and is serving in Game 2. The score for Game 2 is currently 20--15. If the player in blue (IND) smashes out of bounds, TPE wins the point. The new score will be IND 1 20, TPE 0 16.}\\
\texttt{</observation>}\\
\texttt{<answer>}\\
\texttt{The new match score will be IND 1 20, TPE 0 16. The event was held in 2023.}\\
\texttt{</answer>}

\vspace{0.6em}

\textbf{Beacon Answer}\\
\textcolor{blue}{The new match score will be IND 1 20, TPE 0 16. The event was held in 2023. (\checkmark)}

\end{minipage}

\end{tabular}

\caption{\textbf{Hint example 1: A real-world agentic problem.}}
\label{tab:case_4}
\end{table*}
\begin{table*}[t]
\centering
\small

\begin{tabular}{
@{}p{0.38\textwidth}
@{\hspace{0.03\textwidth}}
p{0.58\textwidth}@{}
}

\begin{minipage}[t]{\linewidth}
\vspace{0pt}
\centering

\includegraphics[
    width=\linewidth,
    height=3.2cm,
    keepaspectratio
]{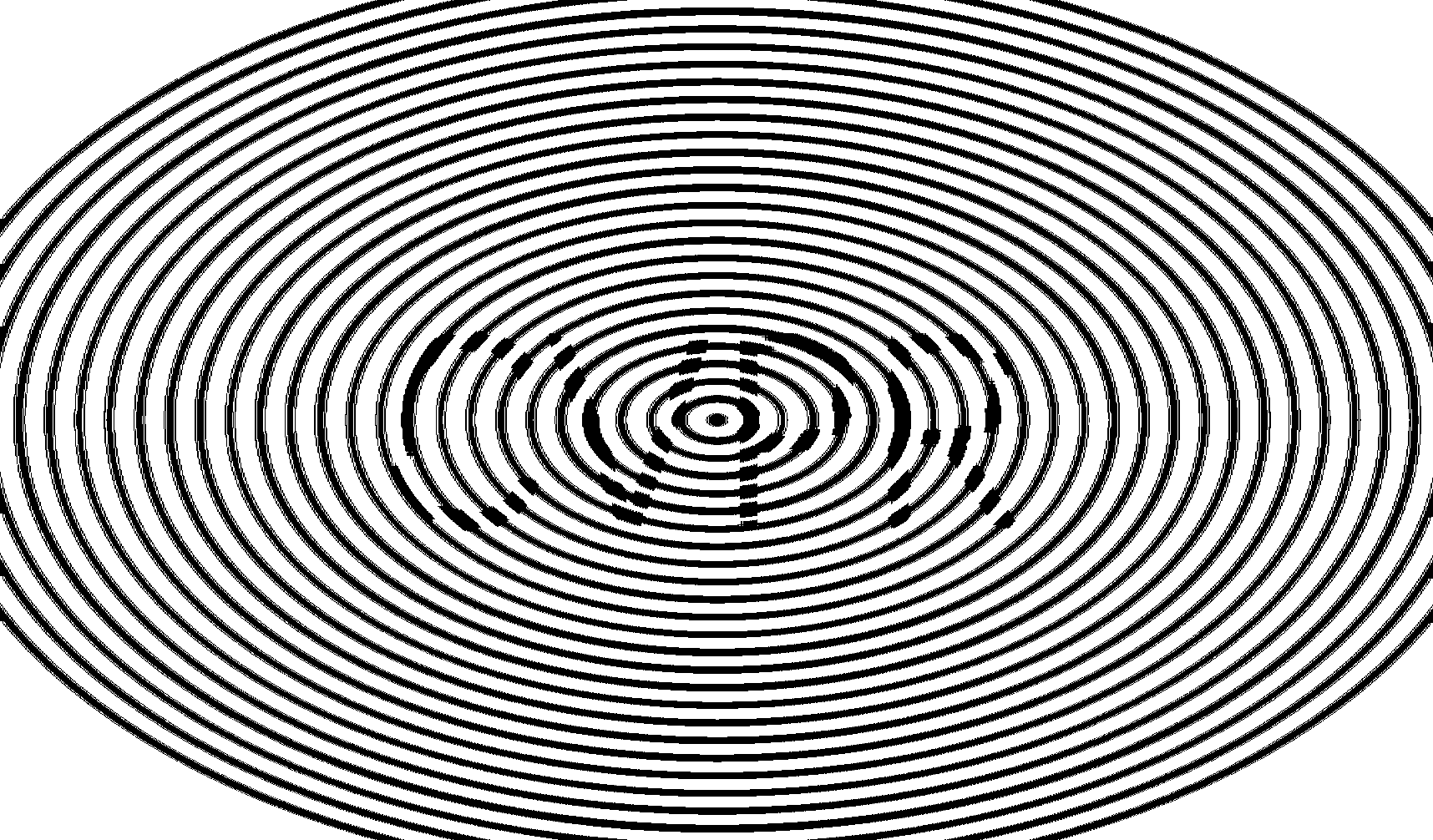}\\[-0.2em]
{\scriptsize Original Image (q0)}

\vspace{0.5em}

\includegraphics[
    width=\linewidth,
    height=3.2cm,
    keepaspectratio
]{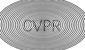}\\[-0.2em]
{\scriptsize g0}

\end{minipage}
&
\begin{minipage}[t]{\linewidth}
\vspace{0pt}

\textbf{Question}\\
Identify all hidden text content in the image

\vspace{0.6em}

\textbf{Hint}\\
1. Hint: [Instruction] Resize the image to a much smaller scale (e.g., 1/10th or 1/20th of the original size); [Expected Subgoal] Reveal any hidden text or patterns that become visible at a lower resolution;

\vspace{0.6em}

\textbf{Beacon Reasoning Given the Hints}\\
\texttt{<tool\_call>}\\
\texttt{img = get\_image("q0")}\\
\texttt{img\_resized = img.resize((img.width // 20, img.height // 20))}\\
\texttt{img\_resized}\\
\texttt{</tool\_call>}\\
\texttt{<tool\_response>}\\
\texttt{generated\_images: g0}\\
\texttt{</tool\_response>}\\
\texttt{\ldots}\\
\texttt{Original image refs: q0}\\
\texttt{Generated image refs: g0}\\
\texttt{</tool\_response>}\\
\texttt{<observation>}\\
\texttt{The resized image clearly shows the text "CVPR" in the center of the spiral pattern. The text was hidden by the high-frequency spiral lines at the original resolution.}\\
\texttt{</observation>}\\
\texttt{<answer>}\\
\texttt{CVPR}\\
\texttt{</answer>}

\vspace{0.6em}

\textbf{Beacon Answer}\\
\textcolor{blue}{CVPR (\checkmark)}

\end{minipage}

\end{tabular}

\caption{\textbf{Hint example 2: Low-resolution OCR.}}
\label{tab:case_5}
\end{table*}
\begin{table*}[t]
\centering
\small

\begin{tabular}{
@{}p{0.38\textwidth}
@{\hspace{0.03\textwidth}}
p{0.58\textwidth}@{}
}

\begin{minipage}[t]{\linewidth}
\vspace{0pt}
\centering

\includegraphics[
    width=\linewidth,
    height=5cm,
    keepaspectratio
]{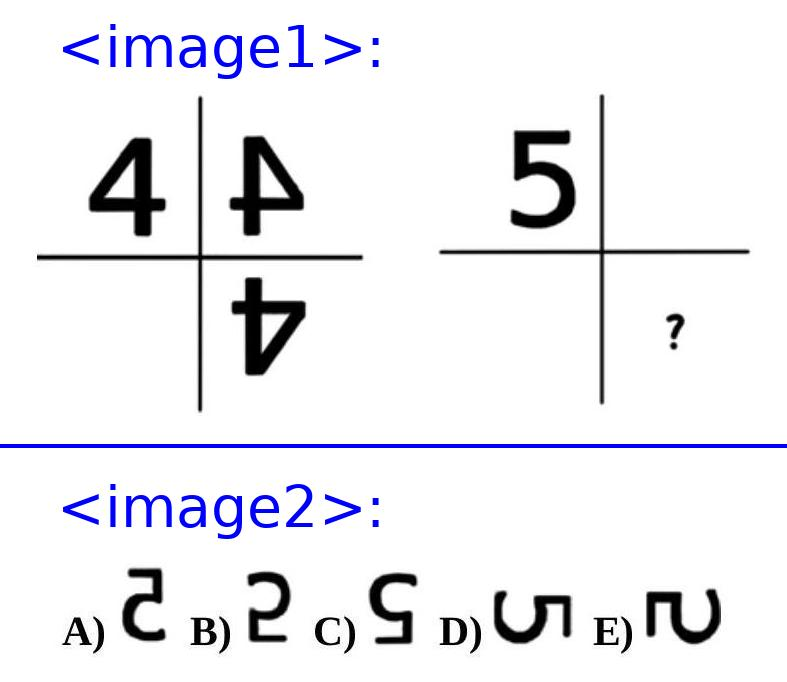}\\[-0.2em]
{\scriptsize Original Image (q0)}

\vspace{0.5em}

\includegraphics[
    width=\linewidth,
    height=3.2cm,
    keepaspectratio
]{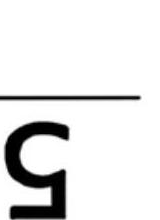}\\[-0.2em]
{\scriptsize g0}

\vspace{0.5em}

\includegraphics[
    width=\linewidth,
    height=3.2cm,
    keepaspectratio
]{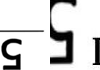}\\[-0.2em]
{\scriptsize g1}

\end{minipage}
&
\begin{minipage}[t]{\linewidth}
\vspace{0pt}

\textbf{Question}\\
The number 4 is reflected twice in the picture. What appears in the field with the question mark if we do the same with the number 5?

\vspace{0.6em}

\textbf{Hint}\\
1. [Instruction] Observe the transformation applied to the number 4, noting that two reflections are equivalent to a 180-degree rotation; [Expected Subgoal] Understand the geometric transformation rule to apply to the number 5;\\
2. [Instruction] Crop the number 5 from the image and rotate it 180 degrees using a tool; [Expected Subgoal] Obtain the visual result of applying the same transformation to the number 5;\\
3. [Instruction] Compare the rotated number 5 with the given options A through E; [Expected Subgoal] Identify which option matches the transformed number 5;

\vspace{0.6em}

\textbf{Beacon Reasoning Given the Hints}\\
\texttt{\ldots}\\
\texttt{img = get\_image("q0")}\\
\texttt{rotated\_5 = img.crop((420, 130, 580, 350)).rotate(180)}\\
\texttt{\# Option C is around x=380 to 450, y=580 to 650}\\
\texttt{opt\_c = img.crop((380, 580, 450, 650))}\\
\texttt{\# Create a side-by-side comparison}\\
\texttt{comp = Image.new('RGB', (100, 70))}\\
\texttt{comp.paste(rotated\_5.resize((50, 70)), (0, 0))}\\
\texttt{comp.paste(opt\_c.resize((50, 70)), (50, 0))}\\
\texttt{comp}\\
\texttt{</tool\_call>}\\
\texttt{<tool\_response>}\\
\texttt{generated\_images: g1}\\
\texttt{</tool\_response>}\\
\texttt{\ldots}\\
\texttt{Original image refs: q0}\\
\texttt{Generated image refs: g0, g1}\\
\texttt{</tool\_response>}\\
\texttt{<observation>}\\
\texttt{The side-by-side comparison shows that the rotated 5 perfectly matches Option C. The top bar is at the bottom, pointing left. The vertical stem is on the right, pointing up. The belly is on the left, curving down and right. This is exactly what happens when you rotate a 5 by 180 degrees. The question asks "What appears in the field with the question mark if we do the same with the number 5?" The transformation is two reflections, which is equivalent to a 180-degree rotation. Therefore, the result is Option C.}\\
\texttt{</observation>}\\
\texttt{<answer>}\\
\texttt{C}\\
\texttt{</answer>}

\vspace{0.6em}

\textbf{Beacon Answer}\\
\textcolor{blue}{C (\checkmark)}

\end{minipage}

\end{tabular}

\caption{\textbf{Hint example 3: Visual puzzles.}}
\label{tab:case_6}
\end{table*}

Tables~\ref{tab:case_4}, \ref{tab:case_5}, and \ref{tab:case_6} demonstrate that training-time hint supervision enables Beacon to extend its reasoning capabilities and tackle more challenging tasks.

\end{document}